\documentclass{article} 
\usepackage{iclr2026_conference,times}
\usepackage[T1]{fontenc}

\usepackage{amsmath,amsfonts,bm}

\def\eqref#1{equation~\ref{#1}}

\def\1{\bm{1}}

\DeclareMathAlphabet{\mathsfit}{\encodingdefault}{\sfdefault}{m}{sl}
\SetMathAlphabet{\mathsfit}{bold}{\encodingdefault}{\sfdefault}{bx}{n}

\usepackage[hidelinks]{hyperref}
\usepackage{url}
\usepackage{graphicx}
\usepackage{amsmath,amssymb}
\usepackage[table]{xcolor}
\usepackage{enumitem}
\usepackage{fancybox}
\usepackage{caption}
\usepackage{booktabs}
\usepackage{longtable}
\usepackage{float}
\usepackage{placeins}
\makeatletter
\let\NewStructureName\@gobble
\let\AssignStructureRole\@gobbletwo
\let\NewTaggingSocket\@gobbletwo
\long\def\NewTaggingSocketPlug#1#2#3{}
\let\AssignTaggingSocketPlug\@gobbletwo
\let\UseTaggingSocket\@gobble

\def\tagstructbegin#1{}

\makeatother
\usepackage{tcolorbox}
\tcbuselibrary{breakable,skins}

\definecolor{scoredeltaup}{RGB}{70,110,170}
\definecolor{scoredeltadown}{RGB}{100,100,110}
\newcommand{\toolup}[1]{{\scriptsize\textcolor{scoredeltaup}{\,{+}#1}}}
\newcommand{\tooldown}[1]{{\scriptsize\textcolor{scoredeltadown}{\,{-}#1}}}
\newcommand{\toolzero}{{\scriptsize\textcolor{scoredeltadown}{\,\(\pm\)0.0}}}

\title{Spatial Competence Benchmark}

\author{
Jash Vira\thanks{Corresponding author.} \\
Independent Researcher \\
\texttt{jashvira2001404@gmail.com}
\And
Ashley Harris \\
Maptek \\
\texttt{ashley.harris@maptek.com.au}
}

\iclrfinalcopy 
\begin{document}

\maketitle

\begin{abstract}
Spatial competence is the quality of maintaining a consistent internal representation of an environment and using it to infer discrete structure and plan actions under constraints.
Prevailing spatial evaluations for large models are limited to probing isolated primitives through 3D transformations or visual question answering.
We introduce the Spatial Competence Benchmark (SCBench), spanning three hierarchical capability buckets whose tasks require executable outputs verified by deterministic checkers or simulator-based evaluators.
On SCBench, three frontier models exhibit monotonically decreasing accuracy up the capability ladder.
Sweeping output-token caps shows that accuracy gains concentrate at low budgets and saturate quickly, and failures are dominated by locally plausible geometry that breaks global constraints.
We release the task generators, verifiers, and visualisation tooling.\footnote{\url{https://github.com/ashleyharris-maptek-com-au/SpatialCompetenceBenchmark/tree/iclr_2026}}
\end{abstract}

\section{Introduction}

Large models have now demonstrated human-level performance in software-engineering and competition mathematics \citep{anthropic2026opus46,qwen2025qwen3,maslej2025aiindex}.
However, certain core aspects of human competence remain largely untested in these models.
Specifically, they lack robust evaluation of the ability to mentally manipulate spatial configurations, understand topological and geometric relationships, and infer actions, usually known as ``spatial intuition''.

Foundational models have made rapid strides on spatial tasks, yet existing benchmarks still use selection-based probes \citep{stogiannidis2025mindgapbenchmarkingspatial,zhang2025spinbenchperspectiverotationlens,li202511plusbenchdemystifyingmultimodalllm,wang2024pictureworththousandwords}, assessing spatial primitives via visual question answering (VQA) or multiple-choice formats.
Benchmarks that target multi-step tasks \citep{li2025unfoldingspatialcognitionevaluating,zhou2025visualizingstepreasoningmira,tang2025legopuzzlesgoodmllmsmultistep,xu2025origamispacebenchmarkingmultimodalllms,spencer2025gamibenchevaluatingspatialreasoning} move closer to real-world applications but typically focus on narrow domains of manipulation rather than planning under global constraints.
Text-only interfaces \citep{guo2026llmspixelsbenchmarkingspatial,rodionov2026floorplanqabenchmarkspatialreasoning} and physics-based grading \citep{huang2025apexempoweringllmsphysicsbased,hu2025lmgamebenchgoodllmsplaying,sun2025probingmechanicalreasoninglarge} each target partial aspects of problems encountered by models in the wild.

Nearly every SCBench task requires a structured executable output (coordinates, edge sets, action sequences) rather than a free-text answer.
Each task is paired with a programmatic verifier or a simulator-based evaluator that awards partial credit for complex tasks.
Certain tasks also include programmatic instance generators that parameterise difficulty and produce arbitrarily large question pools, guarding against memorisation.
We also evaluate three frontier models and analyse accuracy across the capability ladder, token-budget efficiency, and failure modes (\S\ref{sec:experiments}).

\section{Benchmark design}
\label{sec:benchmark}

A hierarchical capability taxonomy organises tasks into three buckets: axiomatic inference, constructive synthesis, and planning.

\paragraph{Axiomatic tasks.}
These tasks require inferring exact discrete structure from formal rules or point sets.
The aim is to strip away distractions and focus on the fundamentals of geometry and topology.
Examples include enumerating guaranteed edges from corner labels or reconstructing adjacency from a recursive bisection tree.

\paragraph{Constructive tasks.}
These tasks require producing geometric objects that satisfy global constraints.
The goal is to allow models to reason about local geometric relationships and construct a globally consistent solution.
Examples include computing the watertight union of intersecting 3D solids or approximating a curved surface with interlocking Lego bricks.

\paragraph{Planning tasks.}
These tasks require long-horizon action sequences under physical or combinatorial constraints.
These tasks are distinguished by sequential state dependence, where each action modifies the environment and constrains subsequent choices, so graders evaluate the terminal state after simulation.
Examples include modifying terrain so rainfall forms target water configurations or sequencing blasts to maximise flat buildable area.

\begin{table}[t]
\centering
\footnotesize
\setlength{\tabcolsep}{5pt}
\caption{Bucket-level accuracy (\%) on SCBench (no tools). Coloured deltas denote tools minus no-tools (percentage points). Per-task breakdown in Table~\ref{tab:per-task-scores} (Appendix~\ref{app:per-task-scores}).}
\label{tab:bucket-summary}
\vspace{2pt}
\begin{tabular}{@{}lrccc@{}}
\hline
\textbf{Bucket} & \textbf{$N$} & \textbf{Claude Sonnet 4.5} & \textbf{Gemini 3 Pro Preview} & \textbf{GPT-5.2} \\
\hline
Axiomatic    &  75 & 49.3 \toolup{6.7} & 81.3 \tooldown{6.7} & 74.7 \tooldown{4.0} \\
Constructive & 171 & 30.2 \toolup{4.3} & 51.4 \toolup{12.3} & 51.9 \toolup{15.0} \\
Planning     &  39 & 27.5 \tooldown{7.8} & 39.0 \toolup{23.6} & 50.0 \tooldown{8.3} \\
\hline
\textbf{Overall} & \textbf{285} & \textbf{34.9 \toolup{3.2}} & \textbf{57.6 \toolup{8.8}} & \textbf{57.6 \toolup{6.8}} \\
\hline
\end{tabular}
\end{table}

\section{Experiments and results}
\label{sec:experiments}

\begin{figure}[H]
\centering
\setlength{\tabcolsep}{6pt}
\begin{tabular}{ccc}
\includegraphics[height=3.2cm]{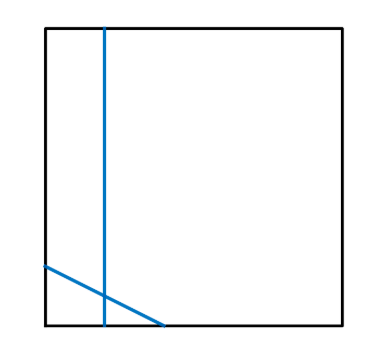} &
\includegraphics[height=3.2cm]{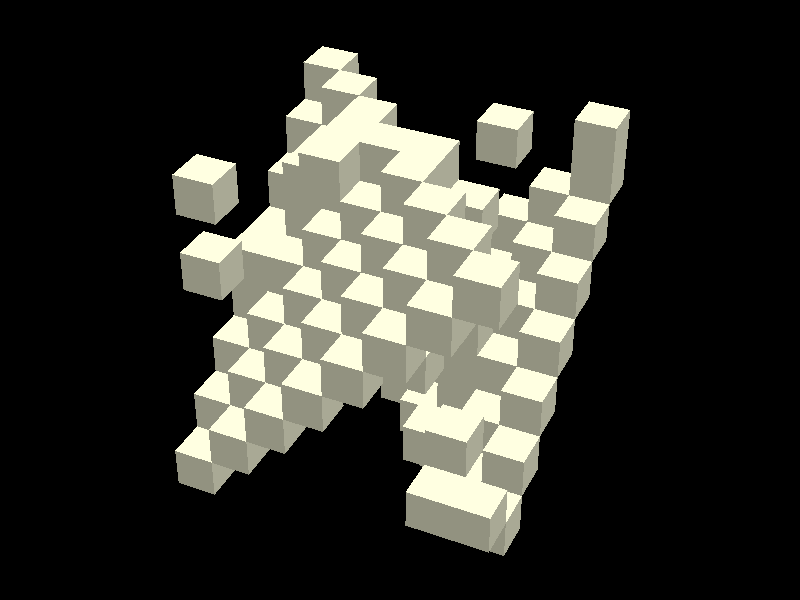} &
\includegraphics[height=3.2cm]{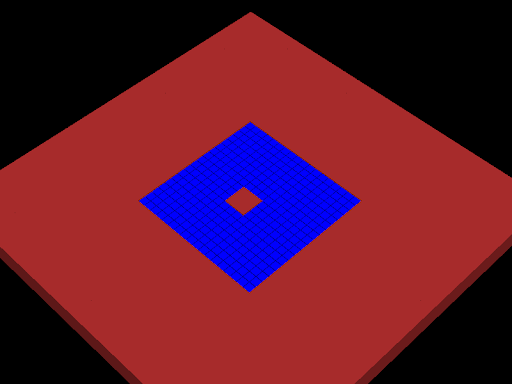} \\[2pt]
{\scriptsize (a) Two Segments (Axiomatic)} &
{\scriptsize (b) Voxel Grid Projection (Constructive)} &
{\scriptsize (c) Fluid Simulation (Planning)} \\
\end{tabular}
\caption{Representative task from each bucket. (a)~Place two segments on a square boundary to partition the interior into a target polygon count; (b)~construct a voxel set in an $N{\times}N{\times}N$ grid whose orthographic projections onto all three axis-aligned planes are fully filled, given constraints; (c)~modify terrain in a 3D voxel world so that rainfall simulation produces a target water configuration.}
\label{fig:task-examples}
\end{figure}

\subsection{Main evaluation}
\label{sec:main-eval}

We evaluate all 22 SCBench tasks (285 subtasks) on three frontier models (Claude Sonnet~4.5, Gemini~3 Pro Preview, GPT-5.2) in two settings (Appendix~\ref{app:eval-protocol}): a \emph{no-tools} setting, where the model receives only the prompt and must emit a schema-conformant artefact scored by a deterministic verifier or simulator, and a \emph{tools} setting, where each model additionally has access to its provider-hosted Python interpreter and web search.
Aggregate results are shown in Table~\ref{tab:bucket-summary}, where coloured deltas denote the tools-minus-no-tools difference in percentage points, with a per-task breakdown in Table~\ref{tab:per-task-scores}.

\textbf{Global constraints are hard to consolidate.}
Gemini and GPT-5.2 tie overall at 57.6\%, both substantially exceeding Sonnet (34.9\%).
The ordering Axiomatic ${>}$ Constructive ${>}$ Planning holds for all three models, suggesting a genuine difficulty gradient rather than model-specific variance.
Within Planning, tasks that admit single-step constructions remain tractable: \emph{Fluid Simulation} (Appendix~\ref{app:fluid-simulation}), where carving a basin traps water, yields strong scores across all models (Sonnet 72.7\%, Gemini 68.2\%, GPT-5.2 100.0\%).
\emph{Terrain Levelling} (Appendix~\ref{app:terrain-leveling}), which requires multi-step dynamics the model cannot query, is unsolved across the board (0\% for all three).

\textbf{Tools help most where computation substitutes for reasoning.}
All three models improve overall with tools (Sonnet ${+}3.2$, Gemini ${+}8.8$, GPT-5.2 ${+}6.8$ pp). The only bucket with consistent gains across all three models is Constructive, where code execution can offload coordinate arithmetic and constraint checking (Sonnet ${+}4.3$, Gemini ${+}12.3$, GPT-5.2 ${+}15.0$), effectively spending attention on global constraints.
On axiomatic tasks, Gemini and GPT-5.2 slightly regress with tools ($-6.7$ and $-4.0$ pp respectively), suggesting that tool overhead can displace effective reasoning on problems already within reach.
Planning shows the widest spread: Gemini gains ${+}23.6$ pp (driven largely by \emph{Hyper-Snake}, ${+}59.3$), while Sonnet and GPT-5.2 both decline.
The largest tools uplift shared across all three models is \emph{Delaunay Triangulation} (Sonnet ${+}56.0$, Gemini ${+}36.0$, GPT-5.2 ${+}48.0$ pp), where a correct library call bypasses the circumcircle reasoning that models consistently fail without tools.

\subsection{Spatial reasoning efficiency}
\label{sec:spatial-reasoning-efficiency}

To measure spatial reasoning efficiency, we use the axiomatic task bucket, where problems are compact and token-efficiency effects are easiest to isolate.

We run no-tools, high-reasoning sweeps for GPT-5.2 and Claude Sonnet~4.5 (further details in Appendix~\ref{app:eval-protocol}), varying the output-token budget $B$.
For GPT-5.2, $B \in \{1024,\;4096,\;8192,\;32{,}768,\;65{,}536\}$.
For Sonnet, $B \in \{1025,\;4096,\;8192,\;32{,}768,\;64{,}000\}$ (provider maximum).
The budget is both enforced by the provider and stated in the prompt, so the model can plan within a known cap.
``Realised output tokens'' denotes the API-reported total generated tokens per instance (including reasoning tokens), averaged across subpasses.

\begin{figure*}[t]
\centering
\begin{minipage}[t]{0.43\textwidth}
\centering
\includegraphics[width=\linewidth]{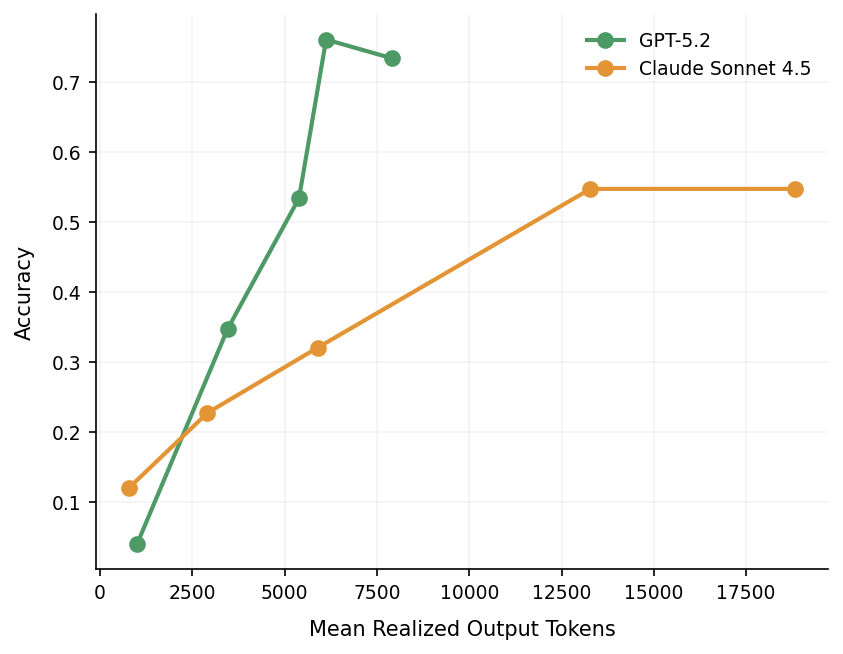}
\caption{Accuracy vs.\ mean realised output tokens across budget caps (highest reasoning mode, no tools).}
\label{fig:budget-pareto}
\end{minipage}
\hfill
\begin{minipage}[t]{0.56\textwidth}
\centering
\includegraphics[width=\linewidth]{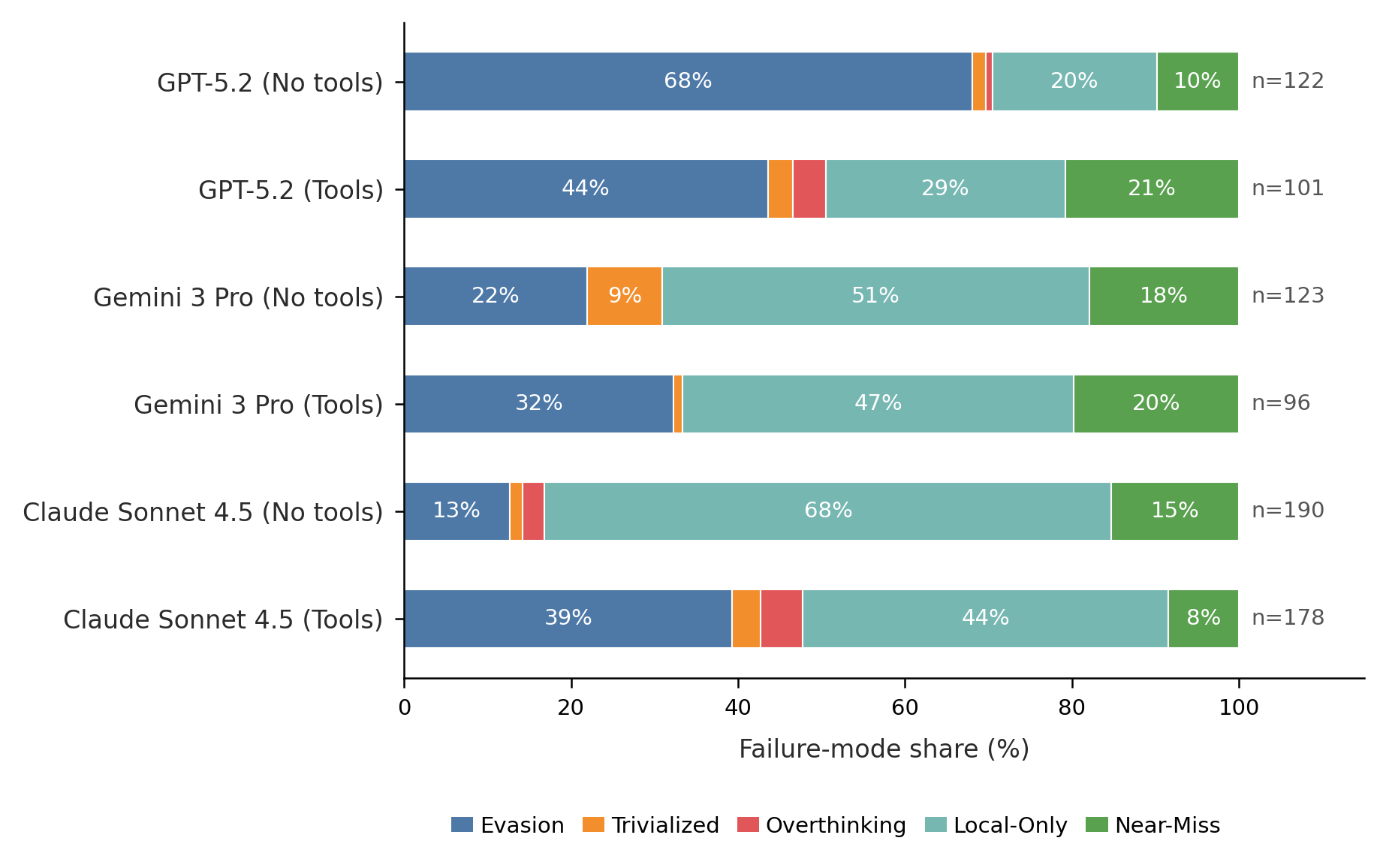}
\caption{Failure-mode distribution over judged failed or low-score subpasses, comparing no-tools and tools runs across all three model families.}
\label{fig:failure-modes}
\end{minipage}
\end{figure*}

Figure~\ref{fig:budget-pareto} shows strong gains through mid budgets and diminishing returns at higher caps for both models.
GPT-5.2 rises from 0.04 at $B=1{,}024$ to 0.76 at $B=32{,}768$, then slightly drops to 0.73 at $B=65{,}536$.
Sonnet rises from 0.12 at $B=1{,}025$ to 0.55 at $B=32{,}768$ and remains flat at $B=64{,}000$.
At comparable budgets, Sonnet consistently consumes more realised output tokens than GPT-5.2 yet reaches a lower accuracy ceiling, indicating that additional tokens do not compensate for less effective spatial reasoning.
At high caps, realised output tokens continue to grow for both models while accuracy plateaus, confirming saturation in spatial reasoning capacity.

\subsection{Failure mode analysis}
\label{sec:failure-modes-text}

Performance scores reveal \emph{that} a model fails but not \emph{why}.
To separate performance measurement from failure attribution, we run a post-hoc diagnostic layer over all subpasses scoring below 0.6.
An independent judge (GPT~5.2-chat with Medium reasoning) receives each failed attempt together with the original prompt, the model's raw output, its chain of thought (where available, as Gemini~3 Pro does not expose reasoning traces), the task's canonical description (see Appendix~\ref{app:task-card-example}), and the programmatic verifier's output, then assigns exactly one label from a fixed set of failure modes.

Forcing a single mutually exclusive label per attempt makes counts comparable across tasks and models; a confidence score flags cases where the available evidence is ambiguous.
We define five mutually exclusive labels:
\begin{description}[nosep,leftmargin=1em,font=\normalfont\bfseries]
\item[Evasion/Forfeit.] Declines the task or returns an empty/placeholder artefact.
\item[Trivialized/Misframed.] Solves a different or easier problem, silently dropping constraints.
\item[Runaway Overthinking.] Extended reasoning that never produces a usable artefact.
\item[Local-Only.] Correct local logic but globally inconsistent solution.
\item[Near-Miss.] Mostly correct artefact with a single verification predicate failing.
\end{description}

\emph{Local-Only} dominates Sonnet and Gemini failure distributions, while GPT-5.2 has a larger \emph{Evasion/Forfeit} share.
In the \emph{Two Segments} (Appendix~\ref{app:two-segments}) task, the model reasons correctly about Euler's formula and face counts, iterates through candidate placements, and produces valid boundary segments, yet the final partition contains a quadrilateral where a pentagon was required: the local geometry is plausible but the global class-count constraint is violated.
A case of \emph{Near-Miss} failures can be seen in \emph{Classify Behaviour} subtask~9 (Appendix~\ref{app:classify-behaviour}), where the model returns a valid label sequence matching the ground truth at every position except one borderline corner configuration, confirming an isolated classification slip rather than systematic failure.
\emph{Evasion/Forfeit} is also frequent across all three models.
On \emph{Delaunay Triangulation} subtask~21 (Appendix~\ref{app:delaunay-triangulation}), the model frames the instance as computationally infeasible, returning an empty triangulation rather than any candidate solution.

Tools shift mass between the failure modes in model-dependent directions (Figure~\ref{fig:failure-modes}).
GPT-5.2's \emph{Evasion} drops from 68\% to 44\% while \emph{Local-Only} rises, Sonnet's \emph{Local-Only} drops from 68\% to 44\% while \emph{Evasion} triples, and Gemini's profile is largely unchanged.
\emph{Local-Only} persists as a dominant residual across all three models, indicating that tools can offload computation but cannot single-handedly close the global-constraint consolidation gap.

\section{Conclusion and Future Work}
\label{sec:conclusion}

We introduced SCBench, a 22-task benchmark spanning axiomatic, constructive, and planning spatial reasoning, graded by deterministic verifiers and physics simulators.
The highest-scoring frontier models attain 57.6\% accuracy, with performance decreasing monotonically from axiomatic to constructive to planning across all three families.
The dominant failure mode is \emph{Local-Only}, where models produce locally plausible geometry but fail to enforce global constraints.
Tools improve scores most on constructive tasks but do not uniformly help across spatial reasoning tasks.
A budget sweep on axiomatic tasks shows most accuracy gains at low token budgets with diminishing returns beyond, and that higher token consumption does not compensate for less effective reasoning, confirming saturation in spatial reasoning capacity.

\paragraph{Future work.}
The current evaluation is limited to single-turn, zero-shot prompts.
Making these evaluations robust across model families, tool-augmented settings, and interaction modes is an important direction.

\newpage
{\small
\bibliography{references}

\begin{thebibliography}{17}
\providecommand{\natexlab}[1]{#1}
\providecommand{\url}[1]{\texttt{#1}}
\expandafter\ifx\csname urlstyle\endcsname\relax
  \providecommand{\doi}[1]{doi: #1}\else
  \providecommand{\doi}{doi: \begingroup \urlstyle{rm}\Url}\fi

\bibitem[{Anthropic}(2026)]{anthropic2026opus46}
{Anthropic}.
\newblock Claude opus 4.6 system card.
\newblock Technical report, Anthropic, 2026.
\newblock URL \url{https://www.anthropic.com/claude-opus-4-6-system-card}.

\bibitem[Guo et~al.(2026)Guo, Yang, Li, Zhang, Gao, Wang, Li, Liu, and Jian]{guo2026llmspixelsbenchmarkingspatial}
Zhongbin Guo, Zhen Yang, Yushan Li, Xinyue Zhang, Wenyu Gao, Jiacheng Wang, Chengzhi Li, Xiangrui Liu, and Ping Jian.
\newblock Can llms see without pixels? benchmarking spatial intelligence from textual descriptions, 2026.
\newblock URL \url{https://arxiv.org/abs/2601.03590}.

\bibitem[Hu et~al.(2026)Hu, Huo, Zhang, Yu, Xing, Stoica, Rosing, Jin, and Zhang]{hu2025lmgamebenchgoodllmsplaying}
Lanxiang Hu, Mingjia Huo, Yuxuan Zhang, Haoyang Yu, Eric~P. Xing, Ion Stoica, Tajana Rosing, Haojian Jin, and Hao Zhang.
\newblock lmgame-bench: How good are llms at playing games?
\newblock In \emph{International Conference on Learning Representations (ICLR)}, 2026.

\bibitem[Huang et~al.(2025)Huang, Yan, Zhang, and Singh]{huang2025apexempoweringllmsphysicsbased}
Wanjing Huang, Weixiang Yan, Zhen Zhang, and Ambuj Singh.
\newblock Apex: Empowering llms with physics-based task planning for real-time insight, 2025.
\newblock URL \url{https://arxiv.org/abs/2505.13921}.

\bibitem[Li et~al.(2025{\natexlab{a}})Li, Wu, Zhang, Li, Gao, Xia, Hernández-Orallo, Vulić, and Wei]{li202511plusbenchdemystifyingmultimodalllm}
Chengzu Li, Wenshan Wu, Huanyu Zhang, Qingtao Li, Zeyu Gao, Yan Xia, José Hernández-Orallo, Ivan Vulić, and Furu Wei.
\newblock 11plus-bench: Demystifying multimodal llm spatial reasoning with cognitive-inspired analysis, 2025{\natexlab{a}}.
\newblock URL \url{https://arxiv.org/abs/2508.20068}.

\bibitem[Li et~al.(2025{\natexlab{b}})Li, Bigverdi, Gu, Ma, Yang, Li, Choi, and Krishna]{li2025unfoldingspatialcognitionevaluating}
Linjie Li, Mahtab Bigverdi, Jiawei Gu, Zixian Ma, Yinuo Yang, Ziang Li, Yejin Choi, and Ranjay Krishna.
\newblock Unfolding spatial cognition: Evaluating multimodal models on visual simulations, 2025{\natexlab{b}}.
\newblock URL \url{https://arxiv.org/abs/2506.04633}.

\bibitem[Maslej et~al.(2025)Maslej, Fattorini, Perrault, Parli, Reuel, Brynjolfsson, Etchemendy, Ligett, Lyons, Manyika, Niebles, Shoham, Wald, and Clark]{maslej2025aiindex}
Nestor Maslej, Loredana Fattorini, Raymond Perrault, Vanessa Parli, Anka Reuel, Erik Brynjolfsson, John Etchemendy, Katrina Ligett, Terah Lyons, James Manyika, Juan~Carlos Niebles, Yoav Shoham, Russell Wald, and Jack Clark.
\newblock The {AI} index 2025 annual report.
\newblock Technical report, Stanford University, Institute for Human-Centered AI, 2025.
\newblock URL \url{https://hai.stanford.edu/ai-index/2025-ai-index-report}.

\bibitem[{Qwen Team}(2025)]{qwen2025qwen3}
{Qwen Team}.
\newblock Qwen3 technical report, 2025.
\newblock URL \url{https://arxiv.org/abs/2505.09388}.

\bibitem[Rodionov et~al.(2026)Rodionov, Eldesokey, Birsak, Femiani, Ghanem, and Wonka]{rodionov2026floorplanqabenchmarkspatialreasoning}
Fedor Rodionov, Abdelrahman Eldesokey, Michael Birsak, John Femiani, Bernard Ghanem, and Peter Wonka.
\newblock Floorplanqa: A benchmark for spatial reasoning in llms using structured representations, 2026.
\newblock URL \url{https://arxiv.org/abs/2507.07644}.

\bibitem[Spencer et~al.(2025)Spencer, Yaari, Vemavarapu, Yang, Ngo, and Sharma]{spencer2025gamibenchevaluatingspatialreasoning}
Ryan Spencer, Roey Yaari, Ritvik Vemavarapu, Joyce Yang, Steven Ngo, and Utkarsh Sharma.
\newblock Gamibench: Evaluating spatial reasoning and 2d-to-3d planning capabilities of mllms with origami folding tasks, 2025.
\newblock URL \url{https://arxiv.org/abs/2512.22207}.

\bibitem[Stogiannidis et~al.(2025)Stogiannidis, McDonagh, and Tsaftaris]{stogiannidis2025mindgapbenchmarkingspatial}
Ilias Stogiannidis, Steven McDonagh, and Sotirios~A. Tsaftaris.
\newblock Mind the gap: Benchmarking spatial reasoning in vision-language models, 2025.
\newblock URL \url{https://arxiv.org/abs/2503.19707}.

\bibitem[Sun et~al.(2025)Sun, Gao, Lyu, Luo, Li, and Deng]{sun2025probingmechanicalreasoninglarge}
Haoran Sun, Qingying Gao, Haiyun Lyu, Dezhi Luo, Yijiang Li, and Hokin Deng.
\newblock Probing mechanical reasoning in large vision language models, 2025.
\newblock URL \url{https://arxiv.org/abs/2410.00318}.

\bibitem[Tang et~al.(2025)Tang, Gao, Zeng, Duan, Sun, Xing, Liu, Lyu, and Chen]{tang2025legopuzzlesgoodmllmsmultistep}
Kexian Tang, Junyao Gao, Yanhong Zeng, Haodong Duan, Yanan Sun, Zhening Xing, Wenran Liu, Kaifeng Lyu, and Kai Chen.
\newblock Lego-puzzles: How good are mllms at multi-step spatial reasoning?, 2025.
\newblock URL \url{https://arxiv.org/abs/2503.19990}.

\bibitem[Wang et~al.(2024)Wang, Ming, Shi, Vineet, Wang, Li, and Joshi]{wang2024pictureworththousandwords}
Jiayu Wang, Yifei Ming, Zhenmei Shi, Vibhav Vineet, Xin Wang, Yixuan Li, and Neel Joshi.
\newblock Is a picture worth a thousand words? delving into spatial reasoning for vision language models.
\newblock In \emph{Advances in Neural Information Processing Systems (NeurIPS)}, 2024.

\bibitem[Xu et~al.(2025)Xu, Lu, Zhao, Tan, Wang, Yuan, Chen, and Xu]{xu2025origamispacebenchmarkingmultimodalllms}
Rui Xu, Dakuan Lu, Zicheng Zhao, Xiaoyu Tan, Xintao Wang, Siyu Yuan, Jiangjie Chen, and Yinghui Xu.
\newblock Origamispace: Benchmarking multimodal llms in multi-step spatial reasoning with mathematical constraints, 2025.
\newblock URL \url{https://arxiv.org/abs/2511.18450}.

\bibitem[Zhang et~al.(2025)Zhang, Corcodel, Hori, Cherian, and Zhao]{zhang2025spinbenchperspectiverotationlens}
Yuyou Zhang, Radu Corcodel, Chiori Hori, Anoop Cherian, and Ding Zhao.
\newblock Spinbench: Perspective and rotation as a lens on spatial reasoning in vlms, 2025.
\newblock URL \url{https://arxiv.org/abs/2509.25390}.

\bibitem[Zhou et~al.(2025)Zhou, Tu, Wang, Wang, Muennighoff, Nie, Choi, Zou, Deng, Yan, Fan, Xie, Yao, and Ye]{zhou2025visualizingstepreasoningmira}
Yiyang Zhou, Haoqin Tu, Zijun Wang, Zeyu Wang, Niklas Muennighoff, Fan Nie, Yejin Choi, James Zou, Chaorui Deng, Shen Yan, Haoqi Fan, Cihang Xie, Huaxiu Yao, and Qinghao Ye.
\newblock When visualizing is the first step to reasoning: Mira, a benchmark for visual chain-of-thought, 2025.
\newblock URL \url{https://arxiv.org/abs/2511.02779}.

\end{thebibliography}
\bibliographystyle{iclr2026_conference}
}

\clearpage
\appendix
\raggedbottom
\definecolor{promptheaderbg}{RGB}{70,110,170}
\definecolor{promptbodybg}{RGB}{237,243,250}
\definecolor{promptborder}{RGB}{140,170,210}
\newtcolorbox{promptbox}[1]{%
  enhanced jigsaw,
  breakable,
  colback=promptbodybg,
  colframe=promptborder,
  boxrule=0.6pt,
  sharp corners,
  left=6pt,
  right=6pt,
  top=4pt,
  bottom=4pt,
  title={#1},
  colbacktitle=promptheaderbg,
  coltitle=white,
  fonttitle=\bfseries\small,
  before upper=\ttfamily\small,
  before skip=\medskipamount,
  after skip=\medskipamount,
  title after break={#1 (continued)}
}

\definecolor{outputheaderbg}{RGB}{100,100,110}
\definecolor{outputbodybg}{RGB}{243,243,245}
\definecolor{outputborder}{RGB}{170,170,180}
\newtcolorbox{outputbox}[1]{%
  enhanced jigsaw,
  breakable,
  colback=outputbodybg,
  colframe=outputborder,
  boxrule=0.6pt,
  sharp corners,
  left=6pt,
  right=6pt,
  top=4pt,
  bottom=4pt,
  title={#1},
  colbacktitle=outputheaderbg,
  coltitle=white,
  fonttitle=\bfseries\small,
  before upper=\ttfamily\small,
  before skip=\medskipamount,
  after skip=\medskipamount,
  title after break={#1 (continued)}
}

\section{Supplementary Results and Experiments Details}
\label{app:supplementary-results}

\subsection{Per-task accuracy}
\label{app:per-task-scores}

\begin{table}[H]
\centering
\footnotesize
\setlength{\tabcolsep}{5pt}
\caption{Per-task accuracy (\%) on SCBench (no tools). Each subtask yields a score in $[0,1]$: eleven tasks use binary pass/fail; the remaining eleven award partial credit (or mixed binary+partial) via task-specific metrics (see individual task appendices). Per-task accuracy is the mean subtask score; bucket and overall accuracy weight every subtask equally. Coloured deltas denote tools minus no-tools (percentage points). $N$ = number of subtasks.}
\label{tab:per-task-scores}
\vspace{2pt}
\begin{tabular}{@{}lrccc@{}}
\hline
\textbf{Task} & \textbf{$N$} & \textbf{Claude Sonnet 4.5} & \textbf{Gemini 3 Pro Preview} & \textbf{GPT-5.2} \\
\hline
\multicolumn{5}{@{}l}{\textit{Axiomatic}} \\
~~Topology Enumeration        & 10 & 40.0 \toolup{20.0} & 100.0 \tooldown{50.0} & 30.0 \toolzero \\
~~Enumerate Edges             &  8 & 12.5 \toolup{50.0} & 75.0 \toolzero & 100.0 \tooldown{12.5} \\
~~Classify Behaviour          & 13 & 38.5 \toolzero & 46.2 \toolzero & 53.8 \tooldown{7.7} \\
~~Half-Subdivision Neighbours & 17 & 58.8 \tooldown{23.5} & 88.2 \toolzero & 82.4 \tooldown{5.9} \\
~~Two Segments                & 27 & 63.0 \toolup{11.1} & 88.9 \toolzero & 88.9 \toolzero \\
\rowcolor{gray!12} Axiomatic subtotal &  75 & 49.3 \toolup{6.7} & 81.3 \tooldown{6.7} & 74.7 \tooldown{4.0} \\
\hline
\multicolumn{5}{@{}l}{\textit{Constructive}} \\
~~Lego Hemispherical Shell    &  8 & 0.0 \toolzero & 0.2 \toolup{15.4} & 0.0 \toolzero \\
~~CSG Union                   & 22 & 40.4 \toolup{0.6} & 67.4 \toolup{4.2} & 83.1 \tooldown{11.5} \\
~~Tetrahedra Shadow Proj.     & 20 & 16.2 \tooldown{7.2} & 31.2 \toolup{20.6} & 4.7 \toolup{68.5} \\
~~Voxel Grid Projection       &  6 & 66.7 \toolzero & 66.7 \toolup{33.3} & 33.3 \toolzero \\
~~Polynomial Curve Fitting    &  9 & 26.2 \tooldown{10.4} & 68.5 \toolup{29.1} & 50.0 \toolup{27.8} \\
~~Delaunay Triangulation      & 25 & 8.0 \toolup{56.0} & 4.0 \toolup{36.0} & 16.0 \toolup{48.0} \\
~~Hamiltonian Loop            & 10 & 50.0 \toolup{10.0} & 70.0 \toolup{10.0} & 80.0 \toolup{10.0} \\
~~Pipe Loop Fitting           & 40 & 35.0 \tooldown{7.5} & 85.0 \tooldown{7.5} & 90.0 \tooldown{5.0} \\
~~Hide and Seek               &  6 & 50.0 \tooldown{16.7} & 66.7 \toolup{16.7} & 100.0 \toolzero \\
~~Pack Rectangular Prisms     &  9 & 68.8 \tooldown{16.0} & 96.3 \tooldown{21.5} & 33.3 \toolup{11.1} \\
~~Interlocking Parts          &  9 & 22.2 \tooldown{22.2} & 0.0 \toolzero & 0.0 \toolzero \\
~~Largest 3D-Printable Prime  &  1 & 0.0 \toolzero & 0.0 \toolzero & 0.0 \toolzero \\
~~Shikaku Rectangles          &  6 & 16.7 \toolup{33.3} & 33.3 \toolup{66.7} & 100.0 \toolzero \\
\rowcolor{gray!12} Constructive subtotal & 171 & 30.2 \toolup{4.3} & 51.4 \toolup{12.3} & 51.9 \toolup{15.0} \\
\hline
\multicolumn{5}{@{}l}{\textit{Planning}} \\
~~3D Maze                     &  6 & 0.0 \toolzero & 50.0 \toolup{16.7} & 16.7 \toolzero \\
~~Hyper-Snake                 & 13 & 21.0 \tooldown{13.2} & 36.3 \toolup{59.3} & 57.7 \tooldown{11.5} \\
~~Fluid Simulation            & 11 & 72.7 \tooldown{12.1} & 68.2 \toolup{4.5} & 100.0 \tooldown{15.9} \\
~~Terrain Levelling           &  9 & 0.0 \toolzero & 0.0 \toolzero & 0.0 \toolzero \\
\rowcolor{gray!12} Planning subtotal &  39 & 27.5 \tooldown{7.8} & 39.0 \toolup{23.6} & 50.0 \tooldown{8.3} \\
\hline
\textbf{Overall}              & \textbf{285} & \textbf{34.9 \toolup{3.2}} & \textbf{57.6 \toolup{8.8}} & \textbf{57.6 \toolup{6.8}} \\
\hline
\end{tabular}
\end{table}

\subsection{Failure-Mode Example}
\label{app:task-card-example}

\begin{figure}[htb]
\centering
\includegraphics[width=\linewidth]{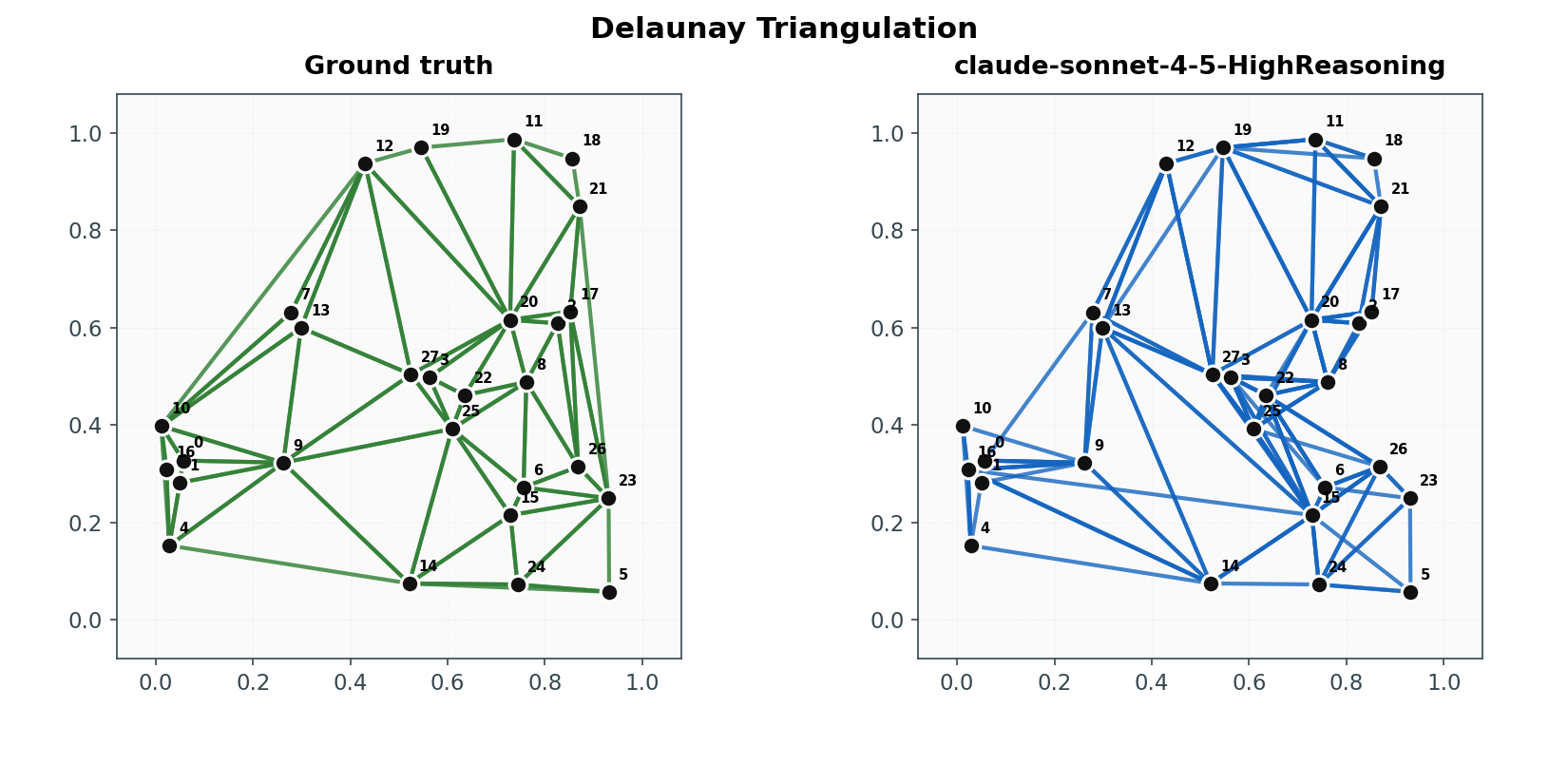}
\caption{Local-Only failure example on \emph{Delaunay Triangulation} (curated 28-point subpass, seed 1009; Claude Sonnet~4.5 no-tools). Left: ground-truth triangulation. Right: model output. Judge label: \textit{Local-Only (Global Constraint Integration Failure)} with confidence 0.90.}
\label{fig:failure-local-only-example}
\end{figure}

For failure-mode analysis (\S\ref{sec:failure-modes-text}), the judge is anchored by a per-task ``card'' that fixes the task intent, output contract, verifier-checked constraints, and tie-break rules. Figure~\ref{fig:failure-local-only-example} shows a concrete Local-Only case from \emph{Delaunay Triangulation} (Appendix~\ref{app:delaunay-triangulation}), using a curated hard subpass with 28 points (seed 1009): the output is schema-conformant, but the verifier reports many missing and extra triangles, indicating local geometric plausibility without globally correct triangulation.

Below is the task card used for this failure-mode classification.

\begin{promptbox}{Task Card example: Delaunay Triangulation}
Task: Delaunay Triangulation\\
INTENT\\
Return the full triangulation for indexed points\\
as a complete set of triangle index triples.\\[6pt]
OUTPUT CONTRACT\\
- Field: triangles (list of 3-integer triples).\\
- Parser recovers from JSON/Python list; canonical key is triangles.\\[6pt]
HARD CONSTRAINTS (verifier-checked)\\
- Every triangle has exactly 3 non-negative indices.\\
- Predicted triangle multiset = ground truth after canonicalisation.\\[6pt]
VERIFIER SIGNALS\\
- passed:False + many missing/extra -> Local-Only\\
- passed:False + one/few mismatches \;\; -> Near-Miss\\
- parse/format failure -> Evasion / Trivialized\\[6pt]
TIE-BREAKS (task-specific)\\
- Evasion/Forfeit: no parseable triangle list.\\
- Trivialized: wrong schema/format despite simple task framing.\\
- Overthinking: long but unstable generation with mixed error signals.\\
- Local-Only: sizeable missing/extra triangle sets.\\
- Near-Miss: only isolated triangle mismatches.\\[6pt]
JUDGE INPUT (inserted into classification prompt)\\
1. Prompt \quad 2. Raw output \quad 3. Parsed triangles\\
4. Verifier diff \quad 5. Reasoning summary (optional)\\[4pt]
>>> Classify with exactly one failure mode.\\
>>> Return JSON: failure\_mode, confidence, justification.
\end{promptbox}

\subsection{Evaluation protocol}
\label{app:eval-protocol}

\begin{table}[H]
\centering
\small
\setlength{\tabcolsep}{6pt}
\begin{tabular}{@{}llll@{}}
\toprule
& \textbf{OpenAI} & \textbf{Anthropic} & \textbf{Google} \\
\midrule
\textbf{Model alias} & \texttt{gpt-5.2} & \texttt{claude-sonnet-4-5} & \texttt{gemini-3-pro-preview} \\
\textbf{Temperature}  & Provider default & Provider default & Provider default \\
\textbf{Reasoning}    & \texttt{effort=xhigh} & \texttt{thinking} enabled (max) & \texttt{thinking\_budget=HIGH} \\
\textbf{Tools}        & \texttt{code\_interpreter}, & \texttt{code\_execution}, & \texttt{code\_execution}, \\
                      & \texttt{web\_search} & \texttt{web\_search} & \texttt{google\_search} \\
\bottomrule
\end{tabular}
\caption{Per-provider evaluation settings for the six runs (single-turn, zero-shot, one sample per instance, no self-correction).}
\label{tab:eval-protocol}
\end{table}

For the spatial reasoning efficiency experiment (\S\ref{sec:spatial-reasoning-efficiency}), we sweep output-token budgets on the axiomatic subset using no-tools, highest-reasoning variants of each model.
$B$ caps total generated output tokens (reasoning $+$ visible text) via each provider's native parameter, and is both enforced provider-side and stated in the prompt.
\begin{itemize}[nosep,leftmargin=1.5em]
  \item \textbf{GPT-5.2:} $B \in \{1024,\;4096,\;8192,\;32768,\;65536\}$
  \item \textbf{Claude Sonnet~4.5:} $B \in \{1025,\;4096,\;8192,\;32768,\;64000\}$.
    The extended-thinking budget is set to $B{-}1$ tokens to allow maximal thinking while following Anthropic's protocol.
\end{itemize}

\section{Benchmark Task Specifications}
\label{app:task-specs}

This appendix provides detailed specifications for each benchmark task in the SCBench. Tasks are subdivided into subtasks, each of increasing complexity.

\subsection{Lego Hemispherical Shell}
\label{app:lego-hemispherical-shell}

The model must construct a hemispherical shell using standard Lego bricks. The goal is to approximate a hemisphere of specified radius while adhering to physical Lego construction constraints including stud alignment, structural stability, and brick interlocking rules.

\begin{figure}[ht]
\centering
\begin{minipage}[t]{0.48\linewidth}
\centering
\includegraphics[width=\linewidth]{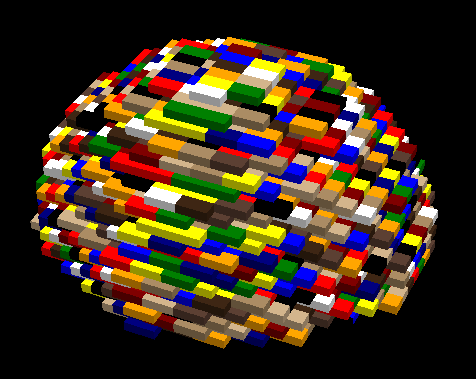}
\end{minipage}
\hfill
\begin{minipage}[t]{0.48\linewidth}
\centering
\includegraphics[width=\linewidth]{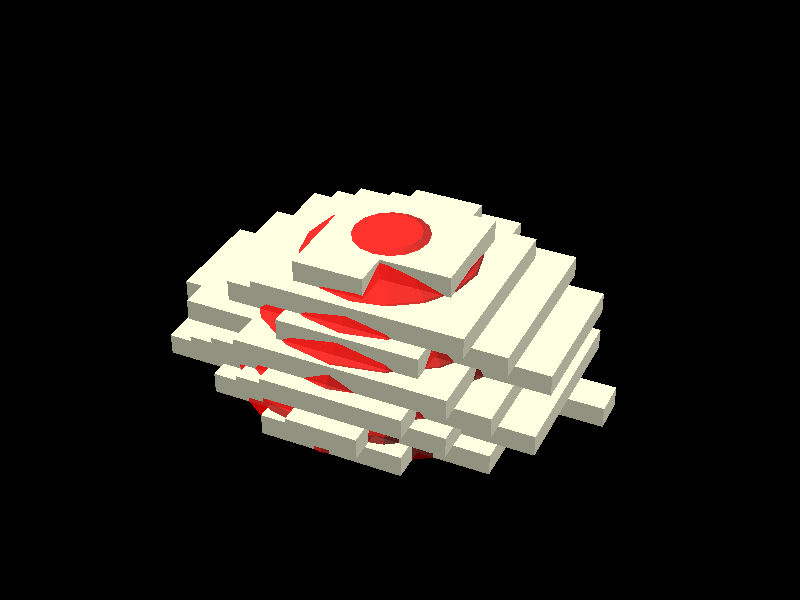}
\end{minipage}
\caption{Lego Hemispherical Shell (1, 2). (Left) A valid Lego hemisphere, 1524 bricks. (Right) Constructive solid geometry (CSG) intersection of Lego structure and hemispherical goal.}
\label{fig:figure-mb2-1-2}
\end{figure}

\begin{promptbox}{Lego Hemispherical Shell -- PROMPT}
You have an unlimited number of Lego(tm) bricks, each of individual size 31.8mm * 15.8mm * 11.4mm but when assembled they are 32mm * 16mm * 9.6mm due to interlocking studs and voids. Each brick has 8 studs, in a 2x4 layout.

These bricks snap together. For example: A brick at (centroid=[0, 0, 4.8],r=90) is resting on the ground centred over the origin, and a second brick at (centroid=[0, 0, 14.4],r=180) is clicked into the first brick with 4 interlocked studs (out of 8). A third brick at (centroid=[24, 0, 24],r=0) is snapped into 2 studs of the second brick, forming a 75\% overhang. Were that third brick at [22,0,24] instead, its shell would intersect 2 studs of the Lego brick below it, and thus not be buildable.

Assemble the bricks such that they resemble a 3D hemispherical shell, with inner radius PARAM\_A cm and outer radius PARAM\_B cm, the centre of the hemisphere is at the origin (0,0,0).

Since it's impossible to create a perfect curve, the best score is one which is closer to the ideal curve, with scoring being calculated based on the volume difference between the ideal curve and the actual brick structure. The structure needs to be buildable in 3D, so bricks can not overlap or be floating in mid air. A great answer does not contain any holes or missing bricks.

Blocks are being placed directly on a flat surface, i.e. without a base plate, so those resting on the ground can have any orientation and are not confined to a specific grid structure. Connected blocks must follow legal Lego(tm) connection rules and must not fall over when built. (Two unstable structures leaning on each other will push each other apart and fall over)

Return a list of the bricks. (location in xyz mm relative to the origin and rotation in degrees).
\end{promptbox}

\noindent\textbf{Verification:}

\begin{table}[H]
\centering\small
\begin{tabular}{@{}cp{0.88\linewidth}@{}}
\toprule
\textbf{Step} & \textbf{Check} \\
\midrule
1 & Each brick not on the ground must connect to another brick (up or down) \\
2 & Bricks must align in X/Y to the 8mm stud grid of any blocks they intersect the studs of \\
3 & No two bricks may occupy the same volume \\
4 & Connected bricks must not form a part with a centre of mass outside of the part's ground footprint \\
5 & No bricks can be below ground, entirely within the inner hemisphere, or entirely outside of the outer hemisphere \\
\bottomrule
\end{tabular}
\end{table}

\paragraph{Subtasks and Scoring.}
There are 8 subtasks, with radii ranging from 4cm-7cm to 15cm-18cm.
CSG operations are done in OpenSCAD, and from this a volume similarity to the reference hemisphere shell is calculated. See Figure~\ref{fig:figure-mb2-1-2} (right).
Due to packing efficiency, a volume overlap of 65\% or better is graded as perfect, and scores are renormalized.

\begin{figure}[ht]
\centering
\begin{minipage}[t]{0.48\linewidth}
\centering
\includegraphics[width=\linewidth]{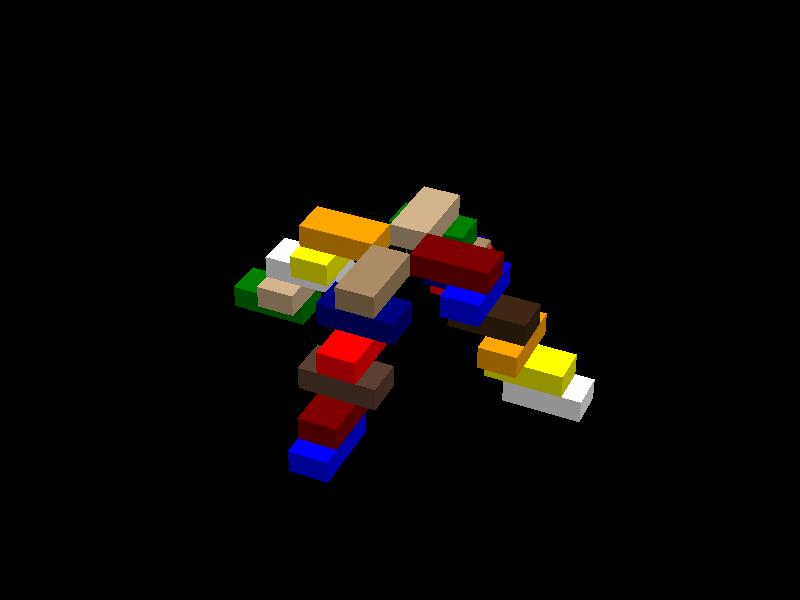}
\end{minipage}
\hfill
\begin{minipage}[t]{0.48\linewidth}
\centering
\includegraphics[width=\linewidth]{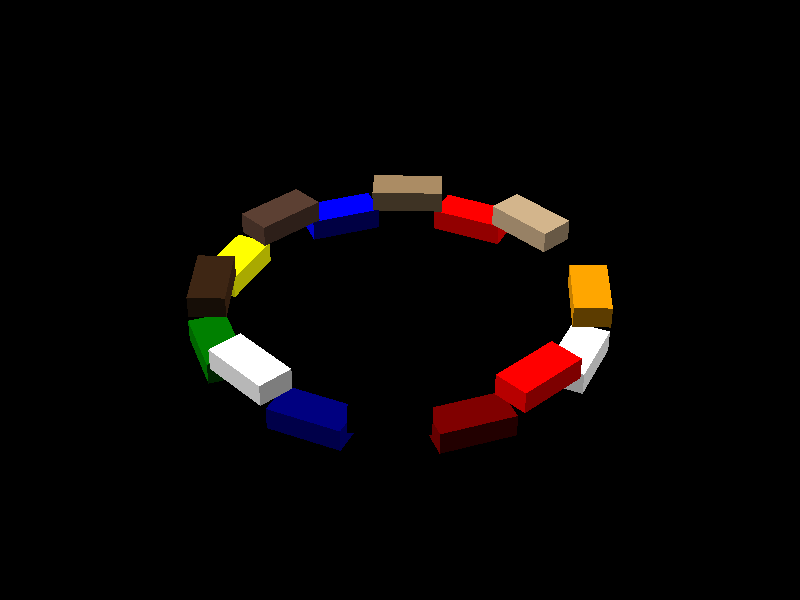}
\end{minipage}
\caption{Lego Hemispherical Shell (3, 4). (Left) An invalid Lego construction will fall over. (Right) Stacked bricks that do not align with the stud grid.}
\label{fig:figure-mb2-3-4}
\end{figure}

\begin{outputbox}{Lego Hemispherical Shell -- OUTPUT FORMAT}
Structured Output is used from the model to mitigate text-parsing or unit errors. A typical output resembles:

[~~ \{'Centroid': [0, 56, 4.8], 'RotationDegrees': 0\}, \{'Centroid': [32, 46, 4.8], 'RotationDegrees': 90\}, \ldots{} ]
\end{outputbox}

\subsection{CSG Union of Polyhedra}
\label{app:csg-union}

The model must compute the constructive solid geometry (CSG) union of multiple 3D primitive shapes and output the resulting polyhedron as a mesh with vertices and faces. The model is given a free-text description of the shapes, must merge them, and convert the result to a structured polyhedron JSON format.

\begin{promptbox}{CSG Union -- PROMPT}
You can output polyhedrons in a json format. For example here is a simple cube, every face has 4 vertices, and there are 6 faces:

\{"polyhedron": \{"vertex":[\{"xyz":[-1.0,-1.0,-1.0]\}, \{"xyz":[1.0,-1.0,-1.0]\}, ...], "faces":[\{"vertex":[3,2,1,0]\}, \{"vertex":[4,5,6,7]\}, ...]\}\}

Now you've learnt the format, use it to solve the following problem:

You are given two cylinders, each with radius 5cm and height 20cm, approximated as 24-sided prisms. The first cylinder has its axis along Z, centered at origin. The second cylinder has its axis along X, centered at origin. They intersect at the center.

Return a polyhedron that is the union of all the solid objects described above. - Ensure faces have normals encoded in a consistent direction (outward-facing). - Ensure no geometry occurs inside the polyhedron, and no faces cross through it. - Note that some faces may have more than 4 vertices. - Your output should have no degenerate faces or redundant vertices. - The task is a failure if the output is not watertight.
\end{promptbox}

\paragraph{Subtasks:}

21 tasks in total, starting with simple cubes, working up to cones, cylinders, spheres, hollowed shapes, and even 3-polyhedra

\begin{figure}[ht]
\centering
\begin{minipage}[t]{0.48\linewidth}
\centering
\includegraphics[width=\linewidth]{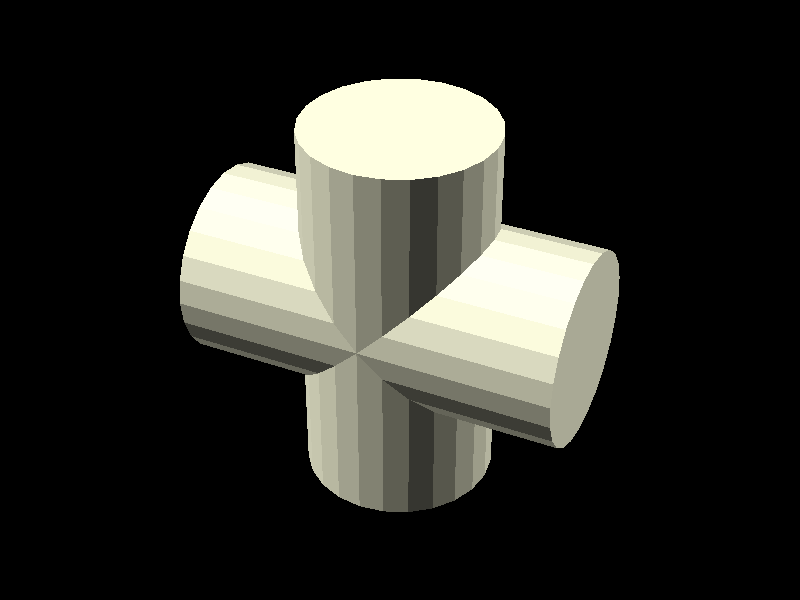}
\end{minipage}
\hfill
\begin{minipage}[t]{0.48\linewidth}
\centering
\includegraphics[width=\linewidth]{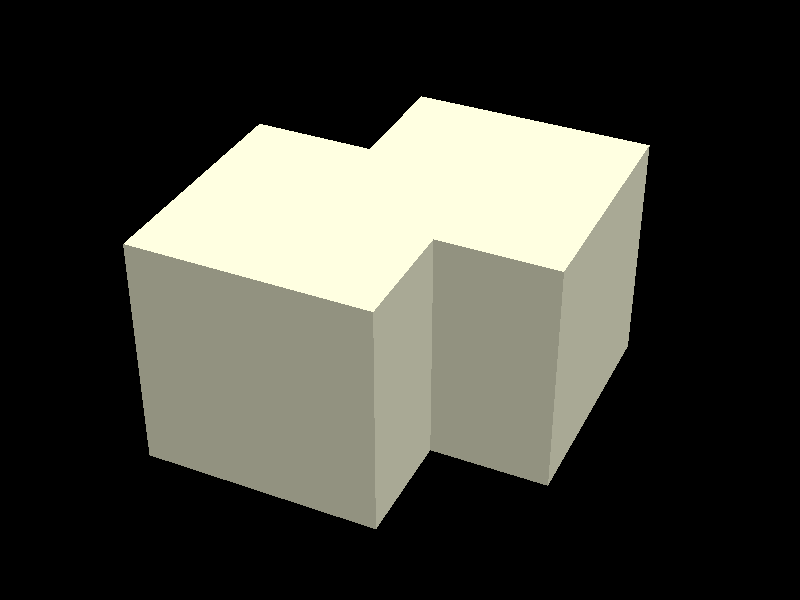}
\end{minipage}
\caption{CSG Union (1, 2). (Left) Two cylinders intersecting at right angles at the origin, as described by the prompt. (Right) Two cubes intersection.}
\label{fig:figure-mb3-1-2}
\end{figure}

\paragraph{Verification}

The returned polyhedron is checked for consistency as follows:

\begin{table}[H]
\centering\small
\begin{tabular}{@{}l@{}}
\toprule
\textbf{Consistency check} \\
\midrule
Empty lists, not 3D, invalid point Ids, NaN positions \\
Degenerate faces and duplicate vertices \\
Edge manifold check \\
Watertightness \\
Self intersecting face check \\
Non-planar face check \\
Winding order check \\
No outlier points ({>}\,100 units away from centroid) \\
Non-zero volume \\
\bottomrule
\end{tabular}
\end{table}

\paragraph{Scoring}

If it passes verification, OpenSCAD is used to calculate the reference CSG, in the cylinder-union example:

\begin{verbatim}
union() {
  cylinder(r=5, h=20, center=true, $fn=24);
  rotate([0,90,0])
    cylinder(r=5, h=20, center=true, $fn=24);
}
\end{verbatim}

This is then used to calculate union, intersection and symmetric differences with the model's provided shape, which are then used to calculate error volume. Above 95\% volume accuracy is considered a perfect score.

\begin{figure}[ht]
\centering
\includegraphics[width=0.55\linewidth]{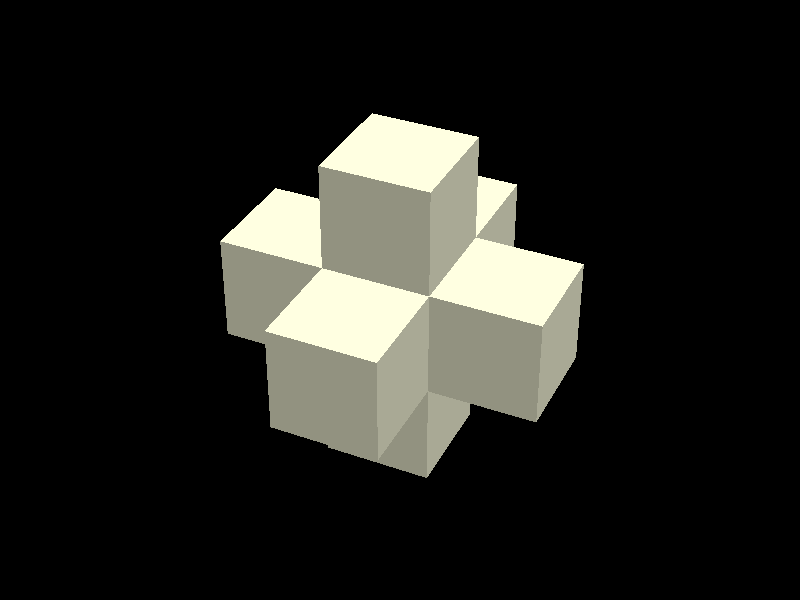}
\caption{CSG Union (3). Three rectangular prism union.}
\label{fig:figure-mb3-3}
\end{figure}

\subsection{Tetrahedra Shadow Projection Coverage}
\label{app:tetrahedra-shadow}

Approximate a target 2D shape (projected shadow) by arranging tetrahedra in 3D space. When viewed as an orthographic projection onto the XY plane, the union of tetrahedra shadows must match the target shape.

\begin{promptbox}{Tetrahedra Shadow Projection -- PROMPT}
Here is the points and faces of a regular tetrahedron with all sides equal 1, resting on the Z=0 plane and with an edge along the X=0 plane:

polyhedron( points = [ [0, 0, 0], [1, 0, 0], [0.5, sqrt(3)/2, 0], [0.5, sqrt(3)/6, sqrt(2/3)] ], faces = [ [0, 2, 1], [0, 1, 3], [1, 2, 3], [2, 0, 3] ]); We define a 7-part rigid transform of (x,y,z,q0,q1,q2,q3), where q0,q1,q2,q3 is a normalised quaternion, and x, y, z are the translation. Rotation is defined around the 0,0,0 point (Which is NOT THE CENTRE), and is performed before translation.

We create a scene with multiple tetrahedra, each with a different transform. Return the transforms of such a scene such that a shadow projected vertically (onto the Z=0 plane) fully covers PARAM, centered at the origin.

Use as many tetrahedra as you need, scoring is based on shadow coverage, not the number of tetrahedra used.

Score will be deducted for: - any shadow outside of the shape - redundant tetrahedra - non-normalised quaternions - intersection tetrahedra
\end{promptbox}

\paragraph{Subtask variations:}

Target 2D shapes include: Square, Circle, Triangle, Hexagon, Star (5-pointed), Cross, Diamond, L-shape, T-shape, Chevron, Arrow, Annulus.

While curves are possible with infinite tetrahedra, we allow a 10-20\% difference in area to avoid penalising excellent finite solutions.

\begin{figure}[ht]
\centering
\includegraphics[width=0.55\linewidth]{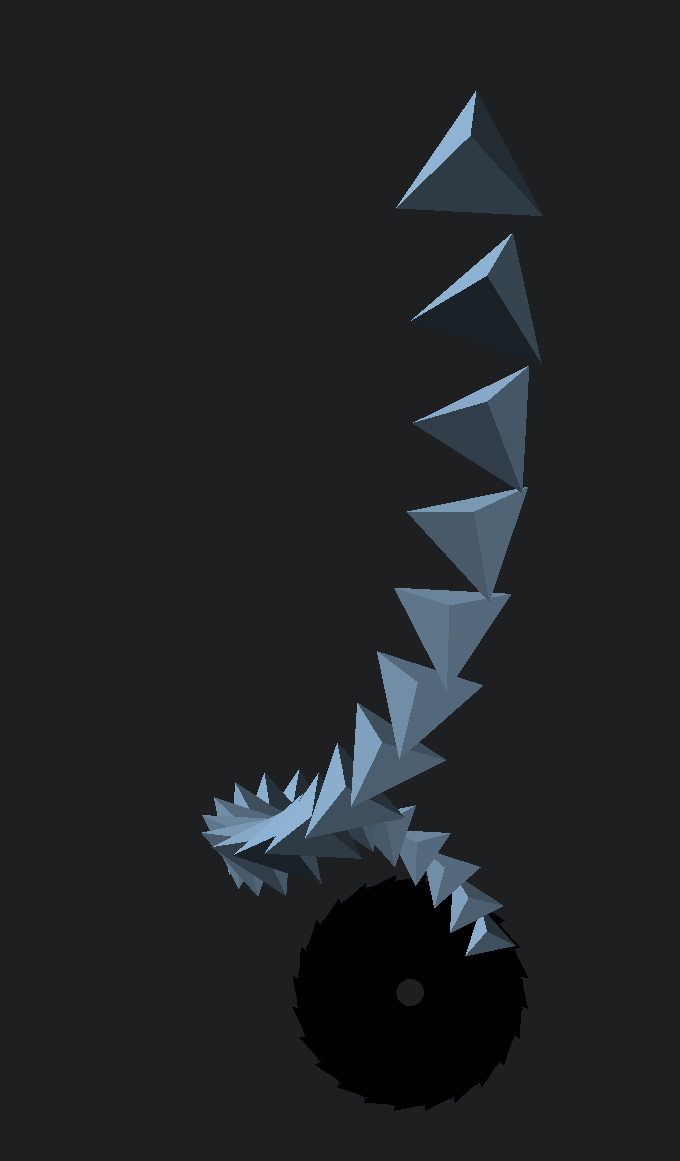}
\caption{The shadow cast from dozens of non-intersecting tetrahedra can form complex 2D shapes, like this washer-shape formed from a spiral.}
\label{fig:figure-mb4-1}
\end{figure}

\noindent\textbf{Verification and scoring:}

\begin{table}[H]
\centering\small
\begin{tabular}{@{}cp{0.82\linewidth}@{}}
\toprule
\textbf{Step} & \textbf{Check} \\
\midrule
1 & Orientation quaternions are checked for normalisation, within 1\% ($-$75\% penalty) \\
2 & Intersection check ($-$50\% penalty) \\
3 & Projection and 2D CSG is performed in OpenSCAD \\
4 & Score calculated from 2D CSG areas of intersection, overlap, and symmetric difference \\
\bottomrule
\end{tabular}
\end{table}

Many models are tripped up by:

\begin{table}[H]
\centering\small
\begin{tabular}{@{}l@{}}
\toprule
\textbf{Common mistake} \\
\midrule
The tetrahedron centre of rotation is not its centre of mass \\
Normalisation of quaternions; this is clearly asked for in the prompt \\
If they try to solve the problem in 2D without considering 3D, the tetrahedra will intersect \\
\bottomrule
\end{tabular}
\end{table}

\paragraph{Solvability:}

Analytic reference shapes (squares, circles, polygons) are generated for each subtask and used directly in verification via CSG intersection and difference volumes in OpenSCAD. Because perfect coverage of curved targets is hard with finite tetrahedra, certain subtasks are renormalised (e.g.\ subpass~1 divides by 0.90).

\subsection{Voxel Grid Projection (Asymmetric Solid)}
\label{app:voxel-grid-projection}

The model must place voxels in a 3D cubic grid subject to two constraints: (1)~all three orthographic projections (XY, XZ, YZ) must be completely solid, and (2)~the voxel arrangement must have no trivial symmetries (rotation or reflection). Each constraint admits a trivial solution in isolation; their conjunction is substantially harder.

\begin{figure}[ht]
\centering
\includegraphics[width=0.55\linewidth]{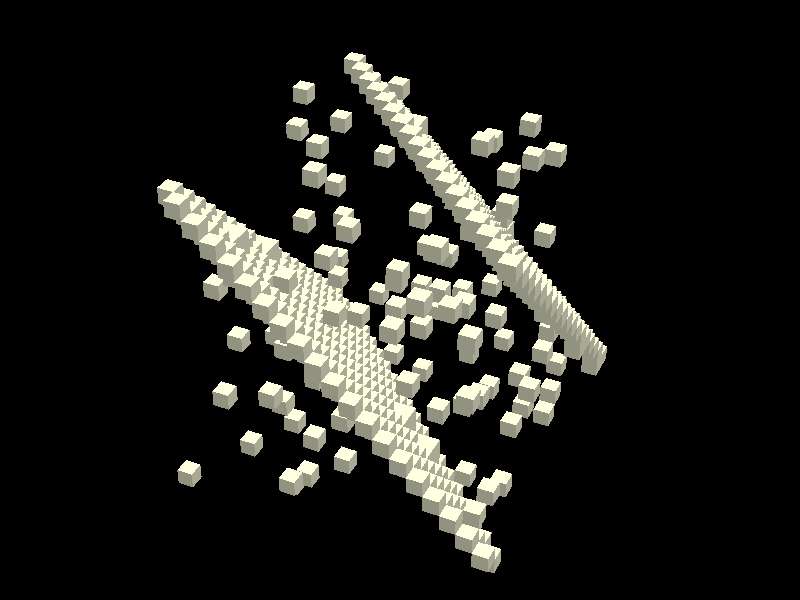}
\caption{Voxel Grid Projection. Valid solution to 20x20x20, 500 voxels, and no voxel having ``7'' in str(x + y + z). ``Noise'' voxels are used to plug gaps and break symmetry.}
\label{fig:figure-mb6-1}
\end{figure}

\paragraph{Subtasks:}
There are 6 subtasks, in which the voxel count and grid size ranges from 6$\times$6$\times$6 with 50 voxels up to 24$\times$24$\times$24 with 1000 voxels, with additional constraints introduced at higher levels.

\begin{promptbox}{Voxel Grid Projection -- PROMPT (EXAMPLE)}
Position 500 voxels in a cubic grid of 20 voxels per side, such that the orthographic projection to all 3 planes is solid (no holes in the projection), and there are no trivial symmetries (rotations or reflections that leave the shape unchanged).

0,0,0 is the bottom left grid cell and indices increase up and to the right.

Ensure no voxels coordinate sum (x + y + z) has a 7 in it.
\end{promptbox}

\begin{outputbox}{Voxel Grid Projection -- OUTPUT SCHEMA}
\begin{verbatim}
{
  "type": "object",
  "properties": {
    "voxels": {
      "type": "array",
      "items": {
        "type": "object",
        "properties": {
          "xyz": {
            "type": "array",
            "items": {
              "type": "number"
            }
          }
        }
      }
    }
  }
}
\end{verbatim}
\end{outputbox}

\noindent\textbf{Verifier Mechanics}

\begin{table}[H]
\centering\small
\begin{tabular}{@{}cp{0.85\linewidth}@{}}
\toprule
\textbf{Step} & \textbf{Check} \\
\midrule
1 & Exact number of voxels required \\
2 & No repeated voxel coordinates \\
3 & All coordinates within [0, N-1] \\
4 & Every (x,y) pair must have at least one voxel at some z \\
5 & Every (x,z) pair must have at least one voxel at some y \\
6 & Every (y,z) pair must have at least one voxel at some x \\
7 & Configuration must not be invariant under any of: 90$^{\circ}$, 180$^{\circ}$, 270$^{\circ}$ rotations about any axis, reflections across any coordinate plane, combinations thereof \\
\bottomrule
\end{tabular}
\end{table}

Pass/fail is binary (all conditions must be satisfied).

\begin{figure}[ht]
\centering
\begin{minipage}[t]{0.48\linewidth}
\centering
\includegraphics[width=\linewidth]{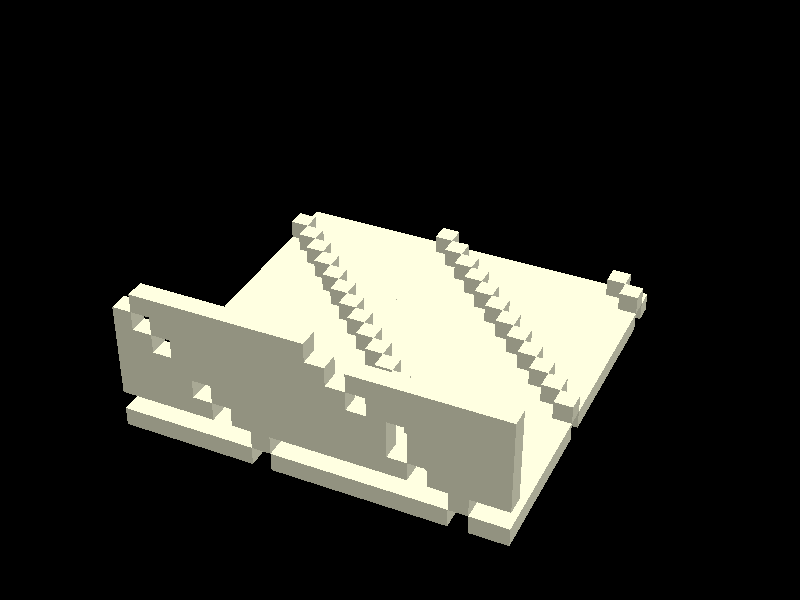}
\end{minipage}
\hfill
\begin{minipage}[t]{0.48\linewidth}
\centering
\includegraphics[width=\linewidth]{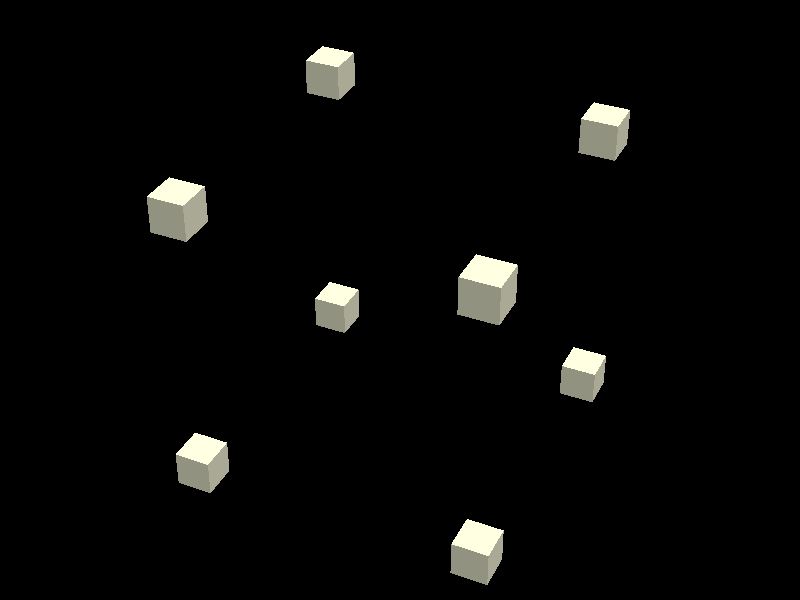}
\end{minipage}
\caption{Voxel Grid Projection (subtasks 3, 4). (Left) The model lost focus while building this structure. Its poor coverage of the XZ and YZ plane scored it a 0. (Right) Poor coverage of a 16x16x16 grid. Voxel coordinates span [4,12] instead of [0,15].}
\label{fig:figure-mb6-3-4}
\end{figure}

\subsection{3D Maze with Jump Mechanics}
\label{app:3d-maze}

The model is asked to design a 3D maze on a heightfield where the solution path requires "jumping" over gaps.

\begin{figure}[ht]
\centering
\includegraphics[width=0.55\linewidth]{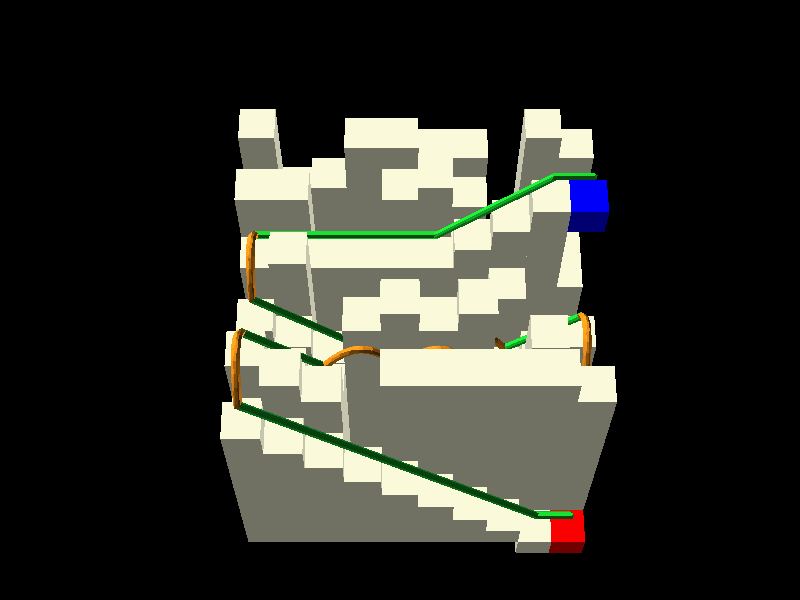}
\caption{3D Maze (1). A valid 3D maze with 7 jumps (shown in yellow).}
\label{fig:figure-mb7-1}
\end{figure}

\paragraph{Maze format}

Output is a raw string: an ASCII grid where each cell contains a character: \texttt{A} (start), \texttt{B} (end), or \texttt{0}--\texttt{9} (elevation, 0 = lowest, 9 = highest). The maze shown above is:

\begin{center}
\includegraphics[width=0.25\linewidth]{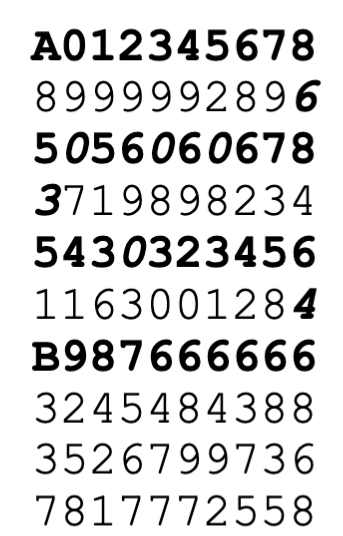}
\end{center}

\noindent\textbf{Rules and restrictions}

\begin{table}[H]
\centering\small
\begin{tabular}{@{}p{0.92\linewidth}@{}}
\toprule
\textbf{Rule} \\
\midrule
Grid must be filled in fully and of size Width $\times$ Height \\
No single elevation value may occupy {>}\,30\% of cells \\
There must only be one solution \\
The path must visit {>}\,20\% of cells \\
You can walk to a neighbouring cell if the elevation is +/$-$ 1 \\
If a neighbouring cell has a drop {$\geq$}\,2, and then there's a cell at the same elevation, you can jump \\
The solution must have a minimum number of jumps \\
Start and End heights are fixed \\
\bottomrule
\end{tabular}
\end{table}

Subpasses vary grid dimensions (5$\times$5 up to 30$\times$30) and minimum number of jumps (2 up to 12). The model must construct both the maze topology and elevation map simultaneously, and ensure that there aren't redundant paths, or loops formed by the ability to take unintended shortcuts. Scoring is binary pass/fail.

\begin{figure}[ht]
\centering
\begin{minipage}[t]{0.48\linewidth}
\centering
\includegraphics[width=\linewidth]{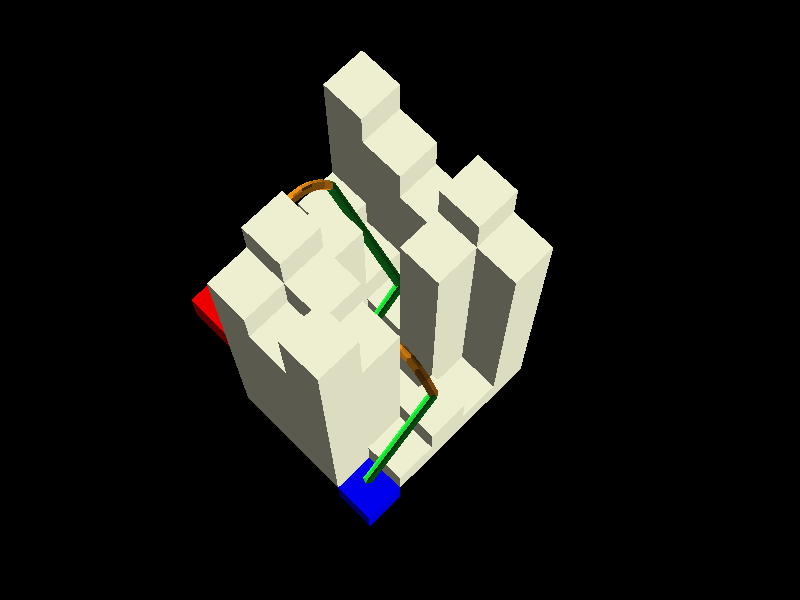}
\end{minipage}
\hfill
\begin{minipage}[t]{0.48\linewidth}
\centering
\includegraphics[width=\linewidth]{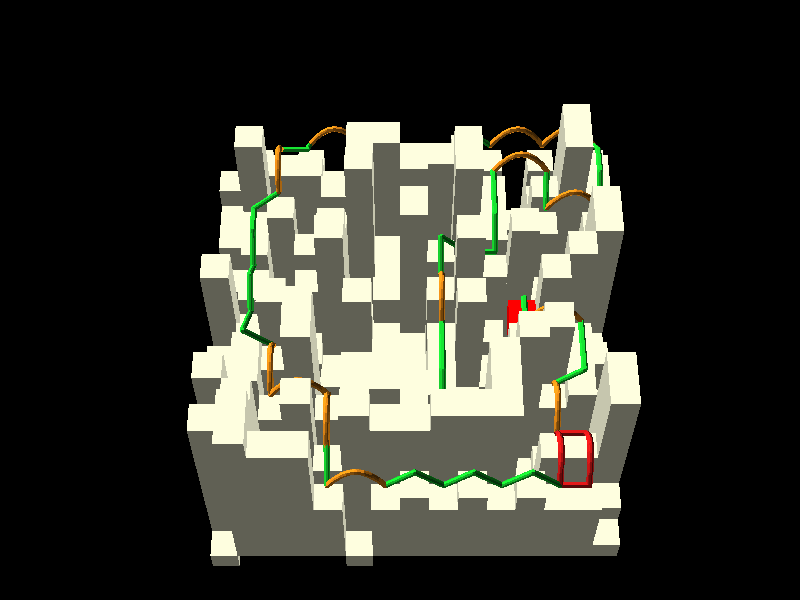}
\end{minipage}
\caption{3D Maze (2, 3). (Left) A 5x5 maze with 2 jumps, a valid solution to subpass 0. (Right) Two parallel jumps (shown in red) mean multiple solutions or loops.}
\label{fig:figure-mb7-2-3}
\end{figure}

\noindent\textbf{Verifier Mechanics}

\begin{table}[H]
\centering\small
\begin{tabular}{@{}cp{0.88\linewidth}@{}}
\toprule
\textbf{Step} & \textbf{Check} \\
\midrule
1 & Trivial things are checked: the grid is square, correctly dimensioned, heights are distributed \\
2 & Breadth-first search (BFS) is used to find the shortest path, including jumps \\
3 & Depth-first search (DFS) is used to find articulation points and discover any `bridges' implying multiple paths \\
\bottomrule
\end{tabular}
\end{table}

\paragraph{Model Performance}

Models without code execution typically produce malformed grids: rows of different lengths, unreachable cells left blank, or missing rows entirely.
Models with code execution usually write an algorithm but forget to verify that their solution accounts for all rules, typically filling unvisitable cells with 9 (making 9 occur too often) or allowing unintended jumps.

\begin{figure}[ht]
\centering
\includegraphics[width=0.55\linewidth]{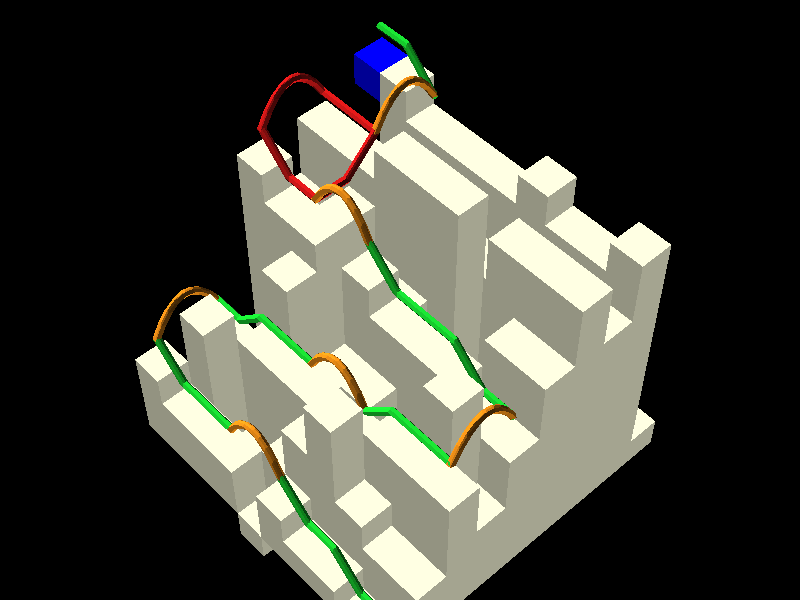}
\caption{3D Maze. A model's attempt at subpass 1, failed due to duplicate paths.}
\label{fig:figure-mb7-4-5}
\end{figure}

\subsection{Polynomial Curve Fitting for 2D Pattern Partitioning}
\label{app:polynomial-curve-fitting}

Given a 2D grid with two distinct regions marked either by different characters in ascii art, (or black and white in images), produce a Python function \texttt{f(x, y)} using only arithmetic operations such that \texttt{f(x,y) > 0} for one region and \texttt{f(x,y) $\leq$ 0} for the other.

\begin{promptbox}{Polynomial Curve Fitting -- TYPICAL PROMPT}
Here is an 12x12 grid representing a space partition:

\begin{verbatim}
#######.....
#######.....
########....
########....
##########..
##########..
##########..
#########...
...####.....
............
............
............
\end{verbatim}

0, 0 is the top left, x is horizontal, y is vertical. Coordinates are in integers.

Using the formula:

let cell = '\#' if f(x,y) > 0 '.' if f(x,y) <= 0

where f(x,y) is a polynomial of whatever degree you need to solve this. You can include cross terms like x*y, x**2*y, x*y**2, etc.

Return the formula as Python function f(x,y) that uses ONLY: - arithmetic operations (+, -, *, /) - powers (**) - parentheses for grouping - integer coordinates x, y - the words "def" and "return"

Do not use type annotations, casts, conditionals, branches, additional variables, comments or anything else.

You can use the following example as a template:

def f(x, y): return x**2 + 3*y**2 - 4*x*y - 145
\end{promptbox}

\paragraph{Subtasks and Scoring.}
There are 9 subtasks. The problem varies from 8x8 up to 128x128. At sizes 48x48 and beyond we switch from ASCII art to monochrome PNG. The patterns are created using Perlin noise and are simple blobs. Polynomial degree ranges from 2 to 10, with coefficient magnitudes increasing across subtasks. Scoring is binary: a subtask is passed if all returned coefficients are within 1\% of the reference coefficients.

\begin{figure}[ht]
\centering
\begin{minipage}[t]{0.48\linewidth}
\centering
\includegraphics[width=\linewidth]{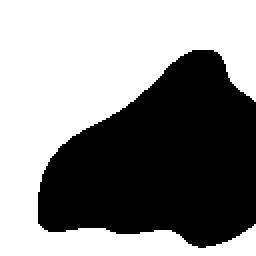}
\end{minipage}
\hfill
\begin{minipage}[t]{0.48\linewidth}
\centering
\includegraphics[width=\linewidth]{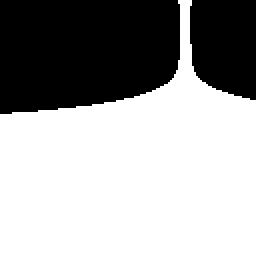}
\end{minipage}
\caption{Polynomial Curve Fitting (2). Expected (left) vs.\ actual (right): a model's attempt at a 128x128 space partition.}
\end{figure}

\noindent\textbf{Verification and grading.}

\begin{table}[H]
\centering\small
\begin{tabular}{@{}cp{0.88\linewidth}@{}}
\toprule
\textbf{Step} & \textbf{Check} \\
\midrule
1 & First the grader checks for forbidden characters. After removing `return', `def' and `f', there must only be x, y, digits, operators, and whitespace \\
2 & Assuming that safety check passes, the code is compiled and executed \\
3 & f(x,y) is evaluated at every grid cell \\
4 & The score is recorded as a fraction of correctly classified cells \\
\bottomrule
\end{tabular}
\end{table}

\subsection{Hamiltonian Loop on Obstructed Grid}
\label{app:hamiltonian-loop}

Find a single closed path that visits every non-blocked cell exactly once on a 2D grid, returning to the starting cell.
Finding a Hamiltonian cycle on a grid graph with obstacles is NP-hard. Small instances are tractable by hand, but na\"ive search algorithms scale poorly.

The model receives:
\begin{itemize}[nosep]
  \item \textbf{Grid dimensions:} Width $\times$ Height
  \item \textbf{Blocked cells:} List of impassable coordinates, presented as an ASCII art map
  \item \textbf{Total valid cells:} Grid area minus blocked cells
\end{itemize}
There are 10 subpasses, ranging from an empty 4x4 to a busy 16x16 grid with many obstacles:

\begin{figure}[ht]
\centering
\includegraphics[width=0.55\linewidth]{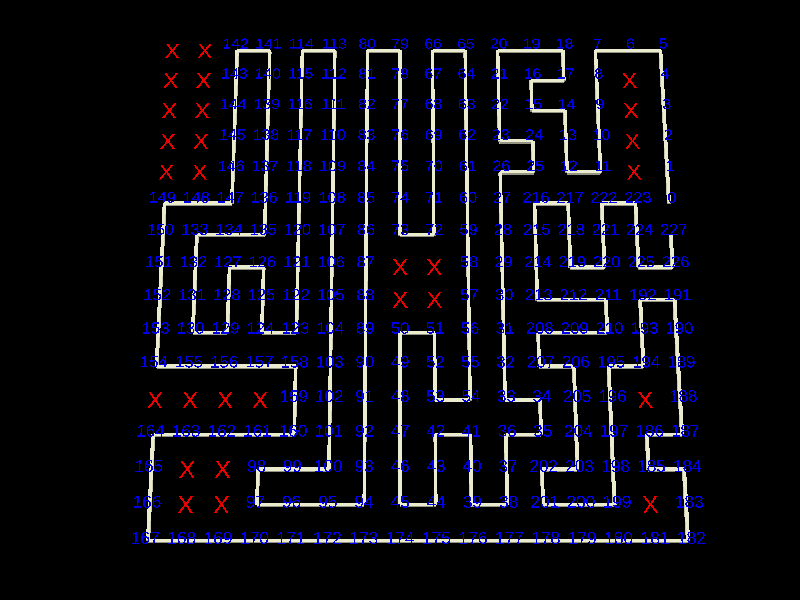}
\caption{Hamiltonian Loop (1). A 16x16 grid with 228 visitable cells and a solved Hamiltonian loop.}
\label{fig:figure-mb9-1}
\end{figure}

\begin{promptbox}{Hamiltonian Loop -- TYPICAL PROMPT}
You have a 4*4 grid of unit squares, with cell coordinates (x, y) where 1 <= x <= 4, 1 <= y <= 4.

.... .X.. .X.. ....

X represents a cell to be skipped. Top Left is 1,1

Draw a single closed path that: - Moves from cell to cell using only side-adjacent moves. - Visits every cell exactly once. - Returns to its starting cell (so the path is a loop). - The last cell in your list must be side-adjacent to the first.

Give an ordered list of the 14 cell coordinates for the loop, starting anywhere.

Structured data is used to ensure that the model returns only a path.
\end{promptbox}

\paragraph{Construction and solvability}

Not all grids are solvable paths, and even fewer are solvable as loops, so problems are only asked of a model if a human-written solver was able to solve the problem.
A human-written algorithm solves this by:

\begin{table}[H]
\centering\small
\begin{tabular}{@{}cp{0.88\linewidth}@{}}
\toprule
\textbf{Step} & \textbf{Action} \\
\midrule
1 & Check a cache of known, solved, subproblems against the grid, seeing if the particular patterns of obstacles have been seen before. Check from largest to smallest. If so, excellent. Mark that chunk as solved \\
2 & Split remaining grid into 2x2 chunks, mark as either solvable (all 4 empty) or unsolvable (has obstacles) \\
3 & Unsolvable chunks grow by absorbing neighbors until solvable \\
4 & Once solvable, attempt to solve using DFS solver. If solved, cache it, otherwise grow \\
5 & Chunks merge when they collide during growth. If chunk size gets too big for DFS, subdivide or restart \\
6 & Once obstacle chunks are solved, grow open space chunks into similar-sized chunks \\
7 & Create adjacency graph of chunks, find Hamilton loop through chunk graph \\
8 & Using that loop, stitch chunks together into one big loop by cutting parallel edges along boundaries \\
\bottomrule
\end{tabular}
\end{table}

\noindent\textbf{Grading}
We use binary pass/fail.

\begin{table}[H]
\centering\small
\begin{tabular}{@{}cp{0.88\linewidth}@{}}
\toprule
\textbf{Step} & \textbf{Check} \\
\midrule
1 & Path length must equal the number of valid (non-blocked) cells \\
2 & All coordinates within bounds of the grid \\
3 & Path may not visit blocked cells \\
4 & Each consecutive pair must be 4-connected / side adjacent \\
5 & Each cell visited exactly once \\
6 & The last cell must be adjacent to the first \\
\bottomrule
\end{tabular}
\end{table}

\subsection{Hyper-Snake Challenge}
\label{app:hyper-snake}

Navigate a "snake" through an N-dimensional grid, avoiding walls and collecting all food items. The snake's body occupies all previously visited cells.

\begin{promptbox}{Hyper-Snake -- TYPICAL PROMPT}
Do you remember the snake game, where you have to direct a snake around a 2D space to avoid hitting itself? This is hyper-snake\!

You are a snake in a 4D space grid of size [5, 5, 5, 5]. You can move to any adjacent cell in the grid, along any of the available 4 dimensions, but you can not move to a cell which you have visited before, nor can you move "diagonally", That is, you can not move in more than one dimension at a time.

The game ends when you run out of space to move, when you hit the boundary, or when you run into yourself.

You need to avoid the obstacles at: [(0, 0, 0, 1), (0, 0, 1, 0), (0, 1, 0, 0)].

You also need to collect the food at: [(0, 0, 0, 0)].

Return the path of the snake as a list of cells, the first element of which is [4, 4, 4, 4] (where you must start), and go for as long as you can.

Structured data is used to ensure that the model's output is a list of Nd points.
\end{promptbox}

\begin{figure}[ht]
\centering
\begin{minipage}[t]{0.48\linewidth}
\centering
\includegraphics[width=\linewidth]{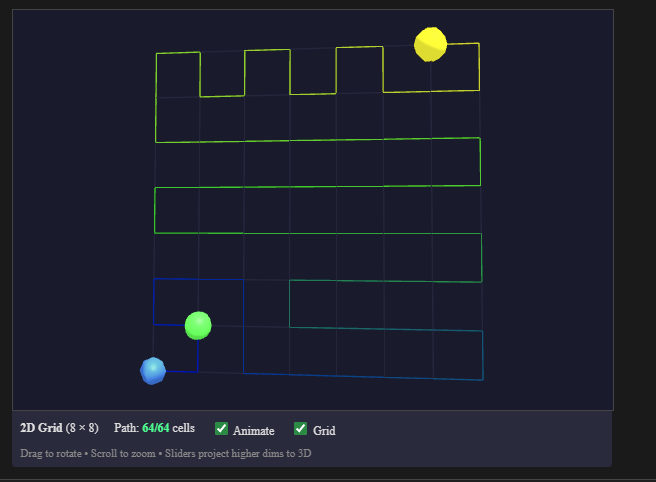}
\end{minipage}
\hfill
\begin{minipage}[t]{0.48\linewidth}
\centering
\includegraphics[width=\linewidth]{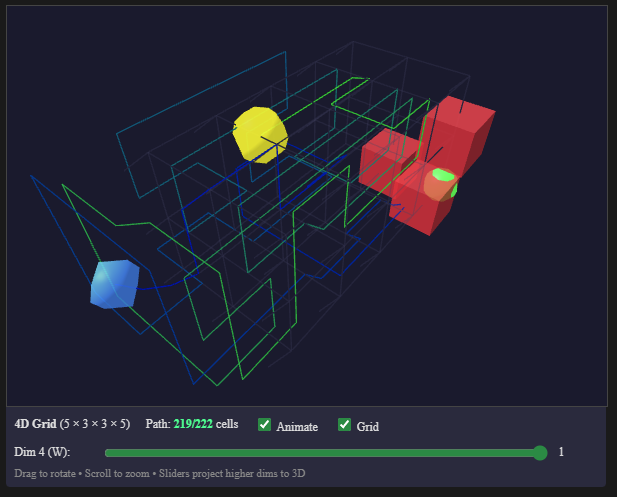}
\end{minipage}
\caption{Hyper-Snake (1). (Left) 2D snake is trivial. The snake (start = blue, head = yellow, body as gradient) collected the food (green) and filled the entire space. (Right) The snake might struggle to access the food due to being blocked by 3 boxes (in X, Y, and Z). However this is a 4D snake, and the W axis is not blocked, allowing the food to be accessed as the snake's final act.}
\label{fig:figure-mb11-1-2}
\end{figure}

\paragraph{Variations and subpasses}

There are 12 subtasks, starting with 2D and ranging up to 10D. Playfield shapes range from 2D grids to simple hypercubes, to an unequal 9D grid of shape $[2, 2, 2, 2, 2, 2, 8, 2, 2]$, to a 10D binary grid. Food, obstacles, and start positions also vary across subtasks to discourage na\"ive space-filling algorithms.

\paragraph{Grading:}

We validate the following:

\begin{table}[H]
\centering\small
\begin{tabular}{@{}cp{0.88\linewidth}@{}}
\toprule
\textbf{Step} & \textbf{Check} \\
\midrule
1 & Path must begin at specified start \\
2 & Each step must have correct number of coordinates \\
3 & Each step changes exactly one coordinate by $\pm$1 \\
4 & All positions within grid extents \\
5 & No position may coincide with a wall \\
6 & Snake body grows; cannot revisit own path \\
7 & All food items must be visited \\
\bottomrule
\end{tabular}
\end{table}

Scoring is calculated as follows:
\begin{itemize}[nosep]
  \item \textbf{Base score:} fraction of visitable cells visited.
  \item \textbf{Food penalty:} uncollected food items reduce the score by a factor of $(1 + n_{\text{missed}})$, where $n_{\text{missed}}$ is the number of uneaten items.
  \item \textbf{Wall adjustment:} when walls are present, the score is divided by 0.98 to compensate for cells made inaccessible by obstacles, scaling 98\% to 100\%.
\end{itemize}

\paragraph{Model performance}

As dimensionality increased, performance declined:

\begin{figure}[ht]
\centering
\begin{minipage}[c]{0.30\linewidth}
\centering\small
\begin{tabular}{@{}ll@{}}
\toprule
\textbf{Dim.} & \textbf{Performance} \\
\midrule
2D & Near-perfect \\
3D & {>}\,60\% \\
4D--6D & 40\%--50\% \\
7D+ & Very poor \\
\bottomrule
\end{tabular}
\end{minipage}
\hfill
\begin{minipage}[c]{0.65\linewidth}
\centering
\includegraphics[width=\linewidth]{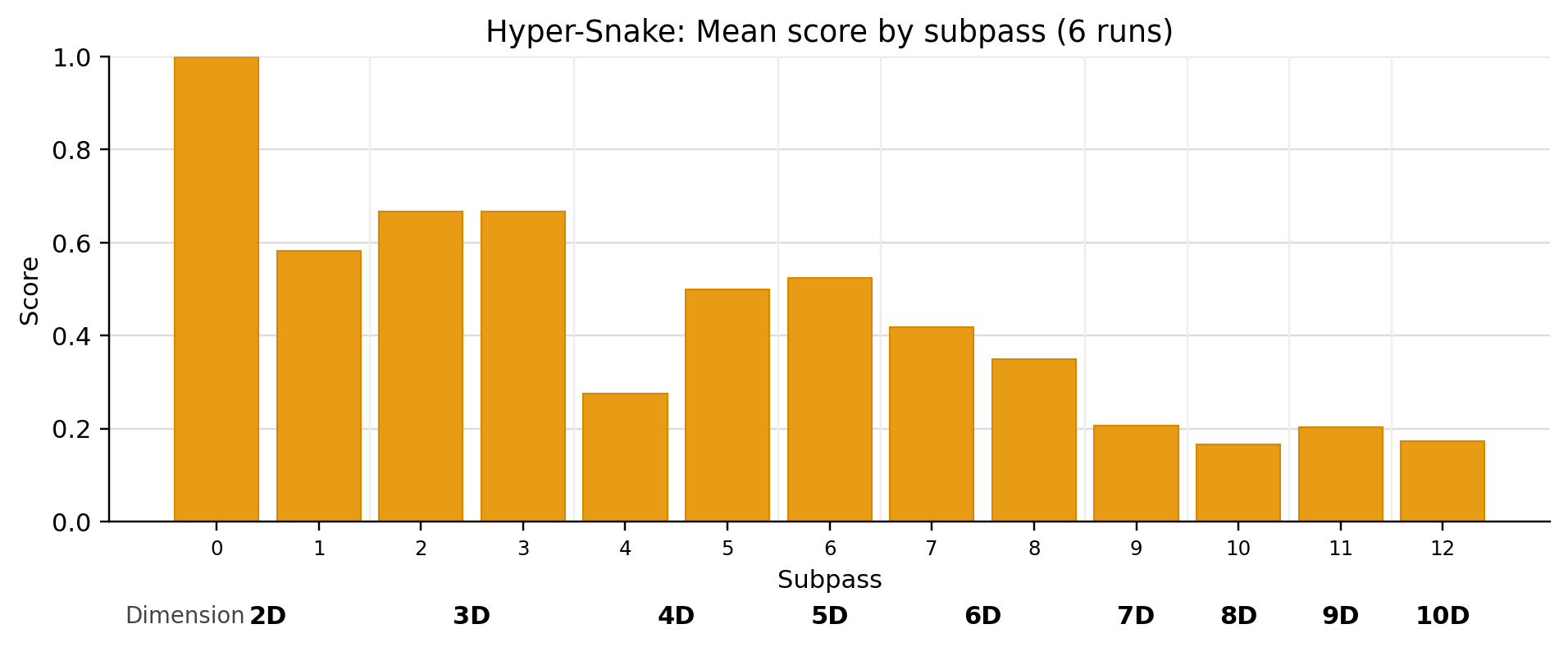}
\end{minipage}
\caption{Hyper-Snake (2). Mean score by subpass across six model runs, with dimensionality progression from 2D to 10D.}
\label{fig:figure-mb11-2}
\end{figure}

\subsection{Pipe Loop Fitting with Constraints}
\label{app:pipe-loop}

Lay out $N$ unit-length pipe segments within a square such that each segment's endpoint touches another, forming a closed loop without self-crossing. Although the total pipe length exceeds the perimeter, the unconstrained problem is trivially solvable and admits infinitely many solutions.

\begin{figure}[ht]
\centering
\includegraphics[width=0.45\linewidth]{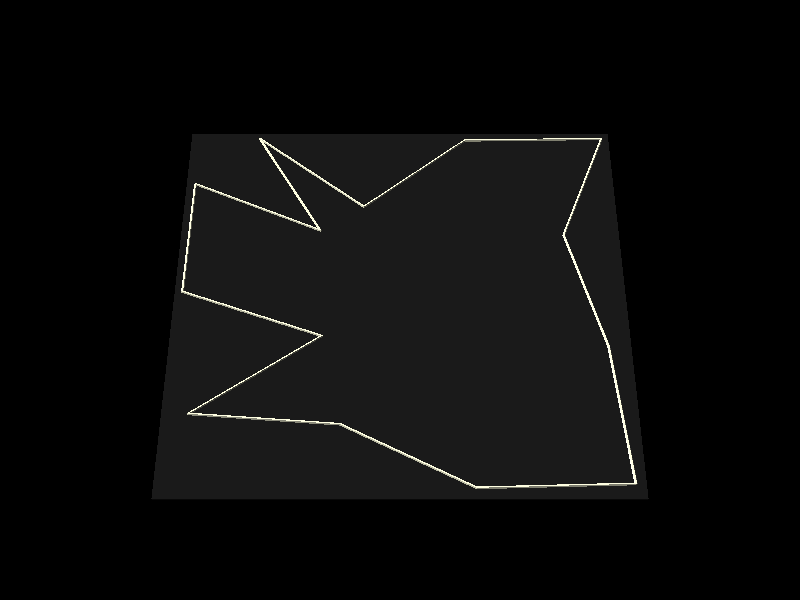}
\caption{Pipe Loop Fitting (1). 14 lengths of pipe laid out end-to-end in a 3x3 square.}
\label{fig:figure-mb12-1}
\end{figure}

\paragraph{Subtasks and Scoring.}

The difficulty increases by adding additional constraints:
\begin{itemize}[nosep]
  \item \textbf{Exact crossings:} loop must self-intersect exactly $N$ times
  \item \textbf{Angle range:} all turn angles must be within $[\theta_{\min}, \theta_{\max}]$ degrees
  \item \textbf{Centroid position:} loop centroid must lie at a specified location
  \item \textbf{Boundary touching:} loop must touch the square boundary at specified edges
  \item \textbf{Minimum turns:} at least $N$ direction changes
  \item \textbf{Maximum straight runs:} no straight segment longer than $L$
  \item \textbf{Quadrant coverage:} loop must pass through all four quadrants
  \item \textbf{Margin avoidance:} stay within an inner margin of the square
  \item \textbf{Minimum convex hull area:} hull of loop points $\geq A$
  \item \textbf{Fixed start point:} first point must be at a specified location
  \item \textbf{Minimum point separation:} all points $\geq D$ apart
\end{itemize}
The square size and pipe count also increase across subtasks, from 3 pipes in a $1{\times}1$ square to 420 pipes in a $50{\times}50$ square. There are 40 subtasks in total.

\begin{promptbox}{Pipe Loop Fitting -- PROMPT EXAMPLE}
You have 36 1m lengths of pipe, and a square area of side length 6 to play with.

Lay the pipe out to form a closed loop, using all the pipe, and returning to the starting point.

You can not re-use vertices, and you can not cross the boundary of the area. You do not need to stick to axis aligned paths.

Return the loop as a list of the pipe endpoints. Note that N pipes require N+1 vertices to describe a path, but since the first and last vertices are the same, you only need to return N points.

Additional constraints: - The center of gravity must stay within 45\%-55\% X and 45\%-55\% Y of the square. - Non-adjacent vertices must be at least 0.25 apart. - Crossings allowed: 0. - The loop must not cross itself or touch itself.

Structured output is used to get an array of 2D vertex coordinates without having to resort to text parsing.
\end{promptbox}

\noindent\textbf{Verifier and grading.}

\begin{table}[H]
\centering\small
\begin{tabular}{@{}ll@{}}
\toprule
\textbf{Stage} & \textbf{Check} \\
\midrule
1.\ Basic validation & Minimum point count \\
  & Loop closure (last point 1m away from first) \\
  & No duplicate vertices \\
  & Within bounds \\
\midrule
2.\ Segment analysis & Compute all line segments \\
  & Detect and count intersections \\
  & Calculate segment lengths and angles \\
  & Check for backtracking \\
\midrule
3.\ Constraint verification & Count intersection points = N \\
  & All turn angles within bounds \\
  & Centroid distance from target {<} tolerance \\
  & Detect edge-touching points \\
  & Count direction changes $\geq$ threshold \\
  & All segment lengths $\leq$ maximum \\
  & Track quadrant visits \\
  & All points within inner box \\
  & Convex hull area $\geq$ minimum \\
  & First point at specified location \\
  & All pairwise distances $\geq$ minimum \\
\bottomrule
\end{tabular}
\end{table}

Binary pass/fail.

\paragraph{Model Performance}

As pipe count and square size increase, performance declines steadily, with poor average scores beyond subtask 35. Subtasks 10--20 carry the bulk of the constraints, producing a visible dip in mean score where simple strategies such as ``distribute around circle'' or ``concentric spirals'' fail (see right panel).

\begin{figure}[ht]
\centering
\begin{minipage}[t]{0.48\linewidth}
\centering
\includegraphics[width=\linewidth]{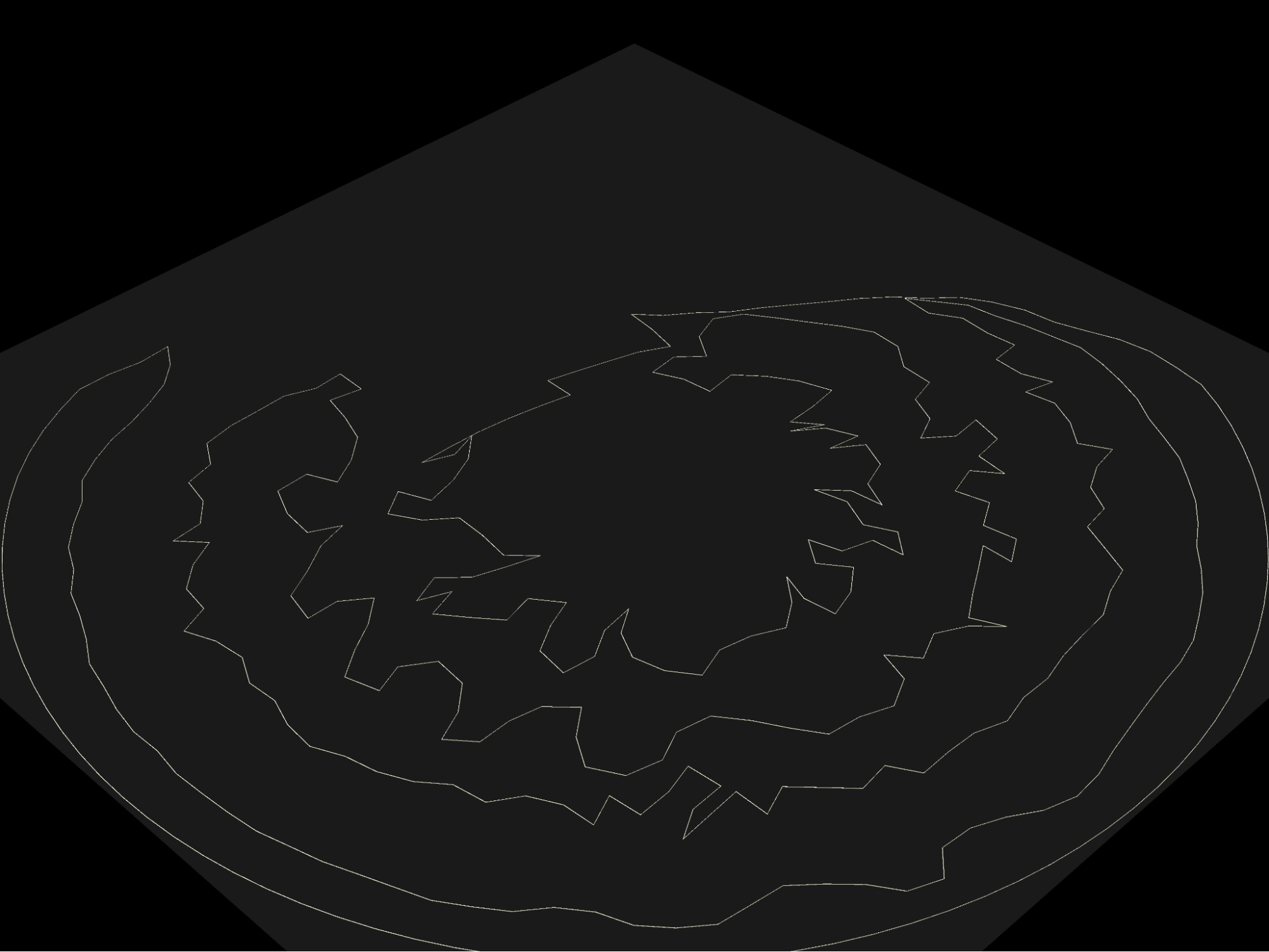}
\end{minipage}
\hfill
\begin{minipage}[t]{0.48\linewidth}
\centering
\includegraphics[width=\linewidth]{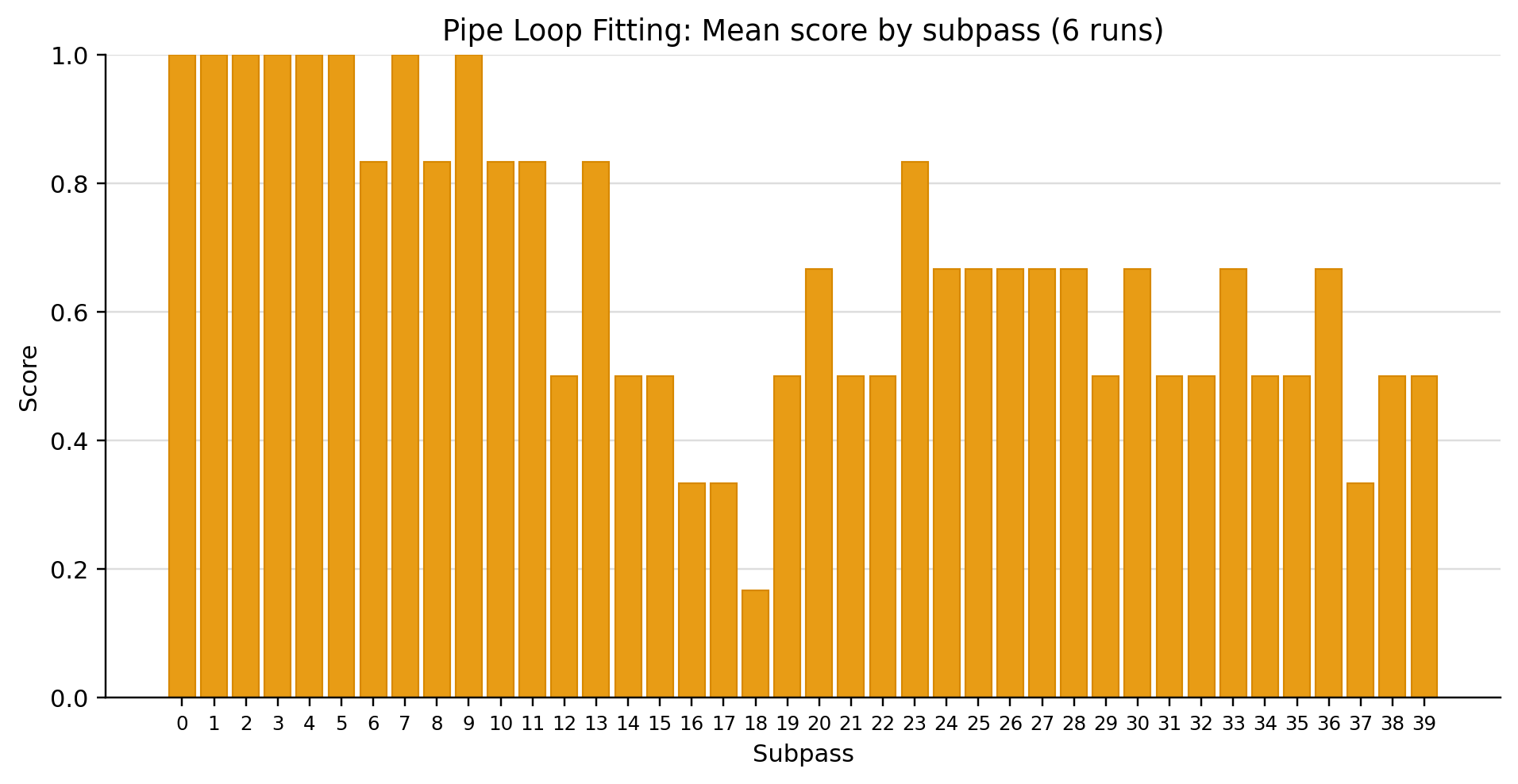}
\end{minipage}
\caption{Pipe Loop Fitting (2, 3). (Left) 390 pipes in 35 x 35, solved with spring solver (subpass 37). (Right) Mean score across all 40 subpasses under six model runs.}
\label{fig:figure-mb12-2-3}
\end{figure}

\subsection{Hide and Seek Behind a Building}
\label{app:hide-and-seek}

Position a crowd of people (represented as axis-aligned bounding boxes) behind a building such that a sniper at a fixed viewpoint cannot see any of them.

\begin{promptbox}{Hide and Seek -- PROMPT TEMPLATE}
You have a building at the origin, axis aligned, PARAM\_B meters wide and deep, and 10 meters tall.

A sniper is located at (100,100,20) and is looking at the building.

Position a crowd of PARAM\_A people (represented by a 0.5*0.5*2m axis aligned bounding box resting on the z=0 plane) in such a way that: - the sniper can not see any of them due to the building blocking their line of sight. - the people must be positioned entirely on the ground (z=0). - the people must not overlap with the building or each other. - nobody is more than 30 meters away from the building's center.

Structured output is used to get an array of x,y coordinates for the location of each person, ensuring the model's answer doesn't need to be parsed from text.
\end{promptbox}

\paragraph{Subpass Parameters}

There are 6 subtasks:

\begin{table}[H]
\centering\small
\begin{tabular}{@{}cll@{}}
\toprule
\textbf{Pass} & \textbf{People} & \textbf{Building width} \\
\midrule
0 & 4 & 2m \\
1 & 20 & 4m \\
2 & 40 & 6m \\
3 & 80 & 8m \\
4 & 150 & 10m \\
5 & 200 & 7m \\
\bottomrule
\end{tabular}
\end{table}

The first few subtasks require only basic geometry (people can be lined up along the diagonal away from the sniper), but this approach fails beyond 40 people. At higher counts the model must reason about projection and line of sight.
A common failure mode is solving the placement for two points, then applying a simple loop without verifying that every person remains occluded.

\begin{figure}[ht]
\centering
\begin{minipage}[t]{0.48\linewidth}
\centering
\includegraphics[width=\linewidth]{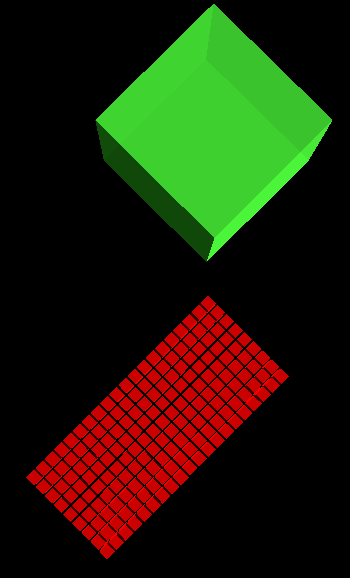}
\end{minipage}
\hfill
\begin{minipage}[t]{0.48\linewidth}
\centering
\includegraphics[width=\linewidth]{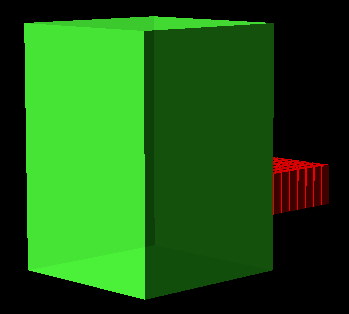}
\end{minipage}
\caption{Hide and Seek (1). View from above the crowd (left) and sniper's view (right). This is a clear failure and scores 0.}
\end{figure}

\paragraph{Grading and validation}

Basic facts are checked first:

\begin{table}[H]
\centering\small
\begin{tabular}{@{}cp{0.88\linewidth}@{}}
\toprule
\textbf{Step} & \textbf{Check} \\
\midrule
1 & Correct number of people \\
2 & All positions are numeric and 2D \\
3 & Each person within maximum distance from building center \\
4 & Person bounding boxes may not intersect building \\
5 & No overlapping bounding boxes \\
\bottomrule
\end{tabular}
\end{table}

A sniper's-view image is then rendered using OpenSCAD and inspected for any visible (red) pixels.
Scoring is not binary: starting from 1.0, a penalty of 0.005 is deducted per visible red pixel (rendered at $800{\times}600$). A fully unobscured person at 25\,m occupies roughly $15{\times}40 = 600$ pixels, so the score is clamped to $[0, 1]$.

\FloatBarrier

\subsection{Pack Rectangular Prisms}
\label{app:pack-rectangular-prisms}

Pack a given set of rectangular prisms into the smallest possible bounding volume. Prisms may be rotated to any orthogonal orientation but may not overlap. This problem is NP-hard.

\begin{figure}[ht]
\centering
\includegraphics[width=0.55\linewidth]{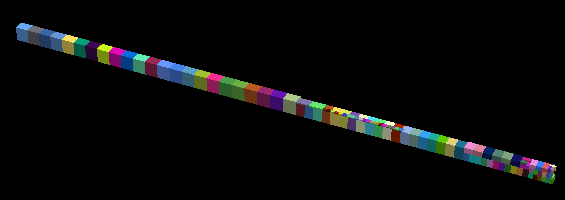}
\caption{Pack Rectangular Prisms (1). 108 prisms (8 size classes) packed 96\% effectively.}
\label{fig:figure-mb16-1}
\end{figure}

\begin{promptbox}{Pack Rectangular Prisms -- TYPICAL PROMPT}
You are given the following rectangular prisms:

7 prisms of 5x3x2 11 prisms of 7x5x3 14 prisms of 11x7x5 17 prisms of 13x11x7 19 prisms of 17x13x11 23 prisms of 19x17x13 15 prisms of 5x5x1 2 prisms of 17x10x2

and have to pack them as efficiently as possible into the smallest volume possible. You can rotate the prisms orthogonally. Return a list of min/max xyz coordinates, one per box that you've successfully packed. A perfect answer is one in which the volume of the enclosing bounding box is the same as the sum of the volumes of the prisms.

Every bit of unused space is a point of failure, effort should be taken to eliminate unused space.

Structured output is used to get the results as schema'd JSON, rather than parse from text.
\end{promptbox}

\paragraph{Solvability}

Not all prism packing instances admit a perfect solution. For example, the instance shown in Figure~\ref{fig:figure-mb16-1} has a total volume of 167{,}593, which cannot be expressed as $x \times y \times z$ with $x, y, z > 1$.

Triple decomposition of the bounding volume allows certain orientations to be pruned from the search space (e.g.\ a $19{\times}17{\times}13$ block can only be placed in two orientations within a $17{\times}17{\times}300$ space), and entire branches can be eliminated when no valid 1D packing of any subset of the allowed sizes spans both boundaries (e.g.\ a span of 16 cannot be constructed from prisms with sides 5, 13, 15, 17).

A human-written solver exploits these insights to reduce complexity. Since the number of size classes is at most 10, perfect solutions can be discovered by checking whether dimension boundaries are achievable with multiples of the given sizes. A greedy algorithm serves as a fallback when a perfect solution is impossible, achieving 95\%+ packing efficiency; this is used as the reference score for grading models on such instances.

\noindent\textbf{Grading process}

\begin{table}[H]
\centering\small
\begin{tabular}{@{}cp{0.88\linewidth}@{}}
\toprule
\textbf{Step} & \textbf{Check} \\
\midrule
1 & Individual box sizes are counted, and we check to make sure that the correct count of each size is present \\
2 & We check for overlaps: no two boxes may intersect or overlap \\
3 & Compute bounding box of all placed boxes. Efficiency = total\_prism\_volume / bounding\_box\_volume \\
4 & If perfect packing is theoretically possible: score = efficiency. Otherwise: score scaled relative to theoretical maximum \\
\bottomrule
\end{tabular}
\end{table}

\paragraph{Model Performance}

Performance drops sharply beyond subtask 2, where the prism count exceeds 20 and exhaustive search becomes infeasible.

\begin{figure}[htb]
\centering
\begin{minipage}[t]{0.48\linewidth}
\centering
\includegraphics[width=\linewidth]{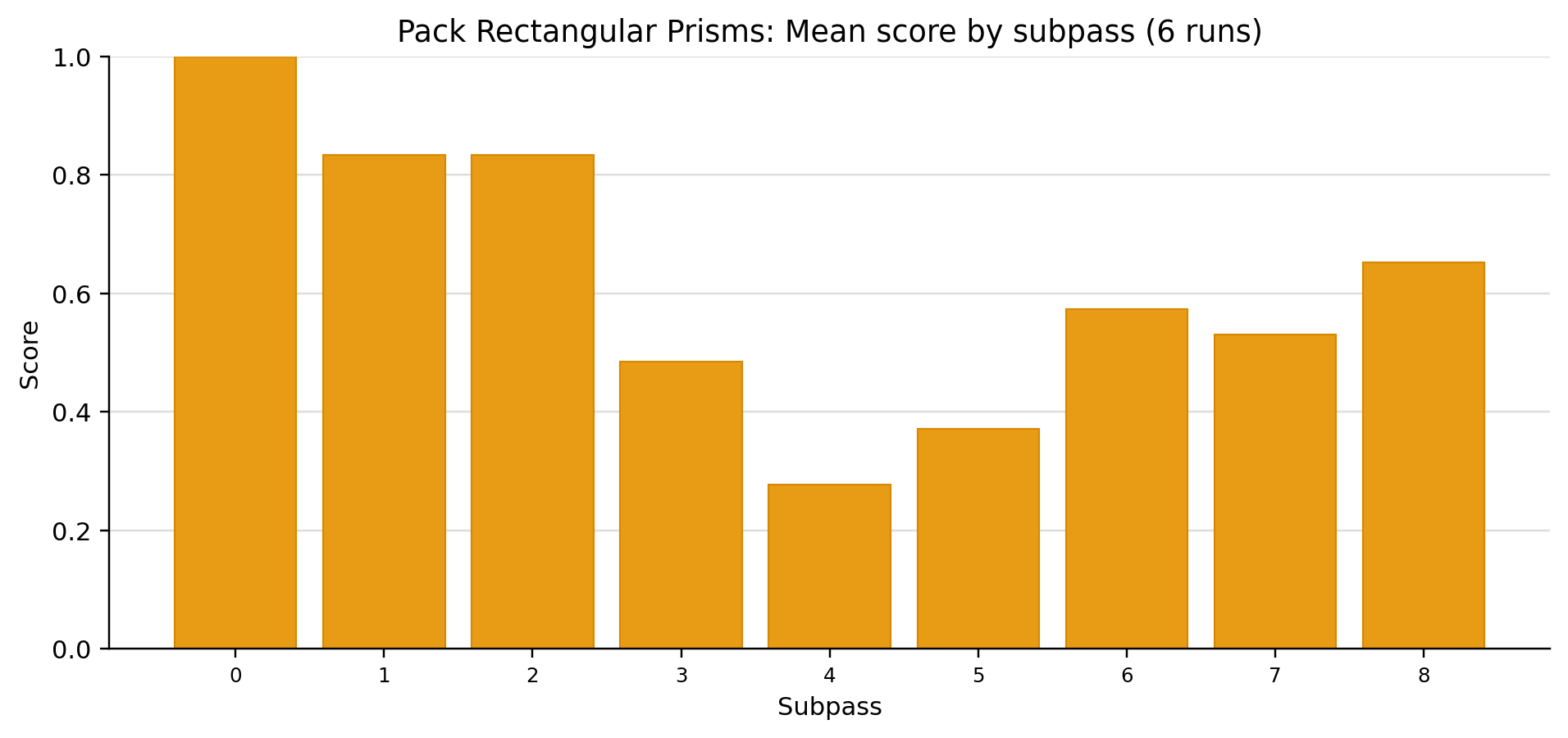}
\end{minipage}
\hfill
\begin{minipage}[t]{0.48\linewidth}
\centering
\includegraphics[width=\linewidth]{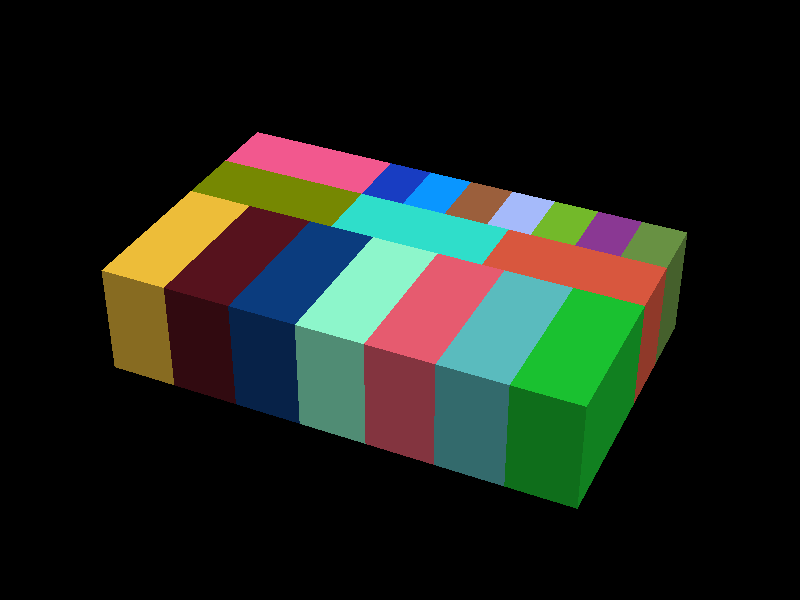}
\end{minipage}
\caption{Pack Rectangular Prisms (2, 3). (Left) Mean score by subtask across six model runs. (Right) A model's perfect packing for subtask 1.}
\label{fig:figure-mb16-2-3}
\end{figure}

\subsection{Fluid Simulation (Earthworks)}
\label{app:fluid-simulation}

Modify a 3D voxel terrain (add/remove rock) such that rainfall results in specific water configurations (lake shapes, volumes, depths).

\begin{figure}[ht]
\centering
\includegraphics[width=0.55\linewidth]{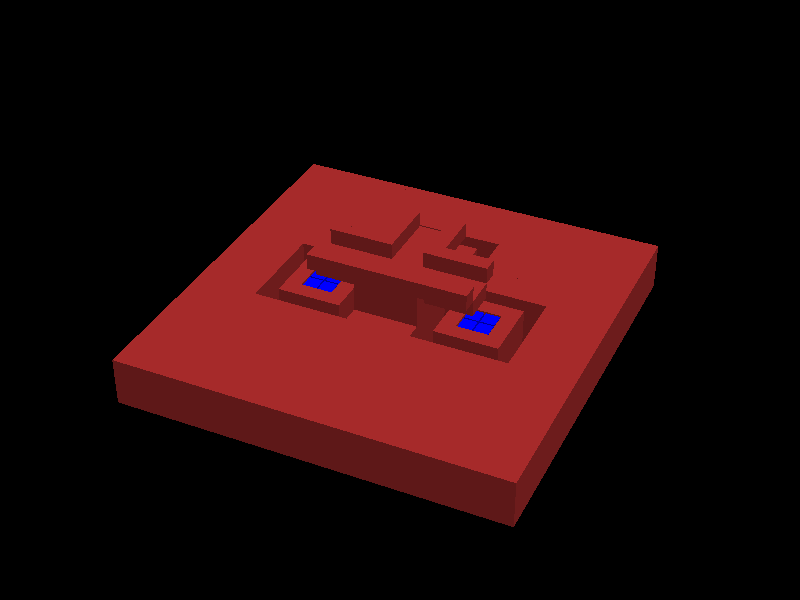}
\caption{Fluid Simulation (1). A successful earthworks project that diverts rainfall from the centre of the map to 3 unconnected lakes.}
\label{fig:figure-mb23-1}
\end{figure}

\begin{promptbox}{Fluid Simulation -- TYPICAL PROMPT}
You are given a 3D voxel world, of dimensions PARAM\_A * PARAM\_A * 8 voxels, currently filled with rock uniformly flat and solid at the layers z = [0,1,2]

Rainfall occurs at the top centre of the world, and follows the following rules: - If air is below water, the water falls (z = z - 1) - If water has ground or water below it, it "flood fills" out looking for any air voxels at the z level below it, and moves to the nearest one. - If water touches the sides or bottom of the world, it falls off and is lost forever. - If water can't find a reachable air voxel to move to, it remains there as a pool of water.

Since the world is currently flat, rainfall currently all runs off the edge of the map, meaning earthworks are required to achieve your goals. After your earthworks complete, there will be 1000s of voxels worth of rainfall (mostly centred at the centre of the world) before your world is graded.

You can specify earthworks using a simple json format: \{ "earthworks" : [ \{ "xyzMin": [2, 2, 3], "xyzMax": [6, 6, 3], "material": "Rock" \}, \{ "xyzMin": [3, 3, 3], "xyzMax": [5, 5, 3], "material": "Air" \} ] \} This would add a 5x5 slab of rock at z = 3, and remove a 3x3 hole from it, leaving a ring of rock at z = 3. This would capture 9 voxels of water.

Rock connected (orthogonally) to other rock is rigid and supports itself, allowing caves, tunnels, bridges, overhangs, and promontories. Floating rock without any support structure will obviously fall to the ground.

Now you understand the format - here is the task I need help with:

Add rock or air voxels to the world in order to ensure that after the rains end, there are 3 lakes on 3 different z levels.

Structured output is used to ensure that the earthworks JSON conforms to schema and is the only output.
\end{promptbox}

\paragraph{Variations:}

There are 11 problems, each requiring a different arrangement of water.

\begin{table}[H]
\centering\small
\begin{tabular}{@{}cp{0.85\linewidth}@{}}
\toprule
\textbf{\#} & \textbf{Requirement} \\
\midrule
1 & A lake of at least 36 voxels in surface area \\
2 & 3 unconnected bodies of water, each at least $2{\times}2$ \\
3 & A lake of at least 2 voxels at $z > 5$ \\
4 & An underground lake of at least 10 voxels in volume but no water voxels are visible from above \\
5 & 3 lakes on 3 different z levels \\
6 & A lake at least 6 voxels deep \\
7 & 2 lakes, each over 200 voxels in volume \\
8 & A ring-shaped lake (water surrounds a dry center of at least $3{\times}3$) \\
9 & Water at $z{=}3$ and $z{=}6$, but no water at $z{=}4$ or $z{=}5$ \\
10 & Exactly 100 voxels of water total \\
11 & 4 separate underground lakes, each at least 5 voxels, none visible from above \\
\bottomrule
\end{tabular}
\end{table}

All problems are verified solvable by a human-written reference answer, none requiring more than 12 earthwork operations.

\noindent\textbf{Grading}

\begin{table}[H]
\centering\small
\begin{tabular}{@{}cp{0.82\linewidth}@{}}
\toprule
\textbf{Step} & \textbf{Check} \\
\midrule
1 & Apply earthworks: modify terrain grid per model instructions \\
2 & Rock stability simulation: floating rock (disconnected) falls until supported \\
3 & Water simulation: drop individual water voxels from above (water falls until hitting rock or other water); flood-fill for pooling behavior; water reaching grid edge drains away; iterate until water stabilizes \\
4 & Evaluation against target: count water surface area; count disconnected water bodies; check water at specific Z levels; check if water is visible from above; measure exact volumes \\
\bottomrule
\end{tabular}
\end{table}

\subsection{Terrain Leveling with Explosives}
\label{app:terrain-leveling}

Plan a sequence of blasting charges on a terrain heightmap to create the largest possible flat area for city construction.

\begin{figure}[ht]
\centering
\begin{minipage}[t]{0.48\linewidth}
\centering
\includegraphics[width=\linewidth]{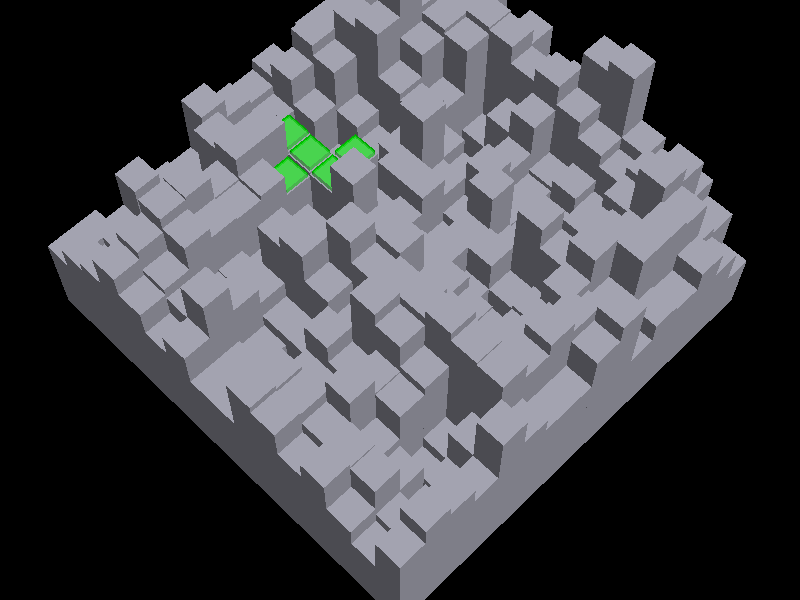}
\end{minipage}
\hfill
\begin{minipage}[t]{0.48\linewidth}
\centering
\includegraphics[width=\linewidth]{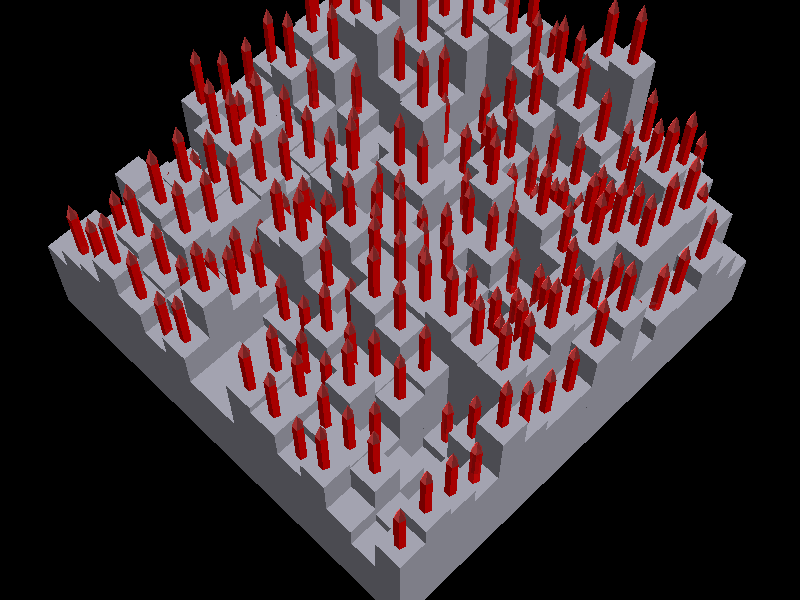}
\end{minipage}
\caption{Terrain Leveling (1, 2). (Left) Typical starting world, with the largest flat city (5 cells) shown in green, 16x16 with z range 1 to 10. (Right) Typical blasting plan.}
\end{figure}

The model receives the heightmap as a 2D array of floats (visualised in the left panel above), along with a worked example of rock fragments rolling downhill and settling in depressions. It must provide a blasting plan (an ordered list of $x, y$ and depth; see right panel). Structured output is used to obtain the blasting plan in JSON form.

The blasting plan is then simulated: rock is fractured and rolls downhill (simulated using PyBullet), settles, and the resulting terrain is written back to the heightmap. The new terrain is checked for the largest flat region, and the score is calculated based on the change in largest flat region size.

\begin{figure}[ht]
\centering
\begin{minipage}[t]{0.48\linewidth}
\centering
\includegraphics[width=\linewidth]{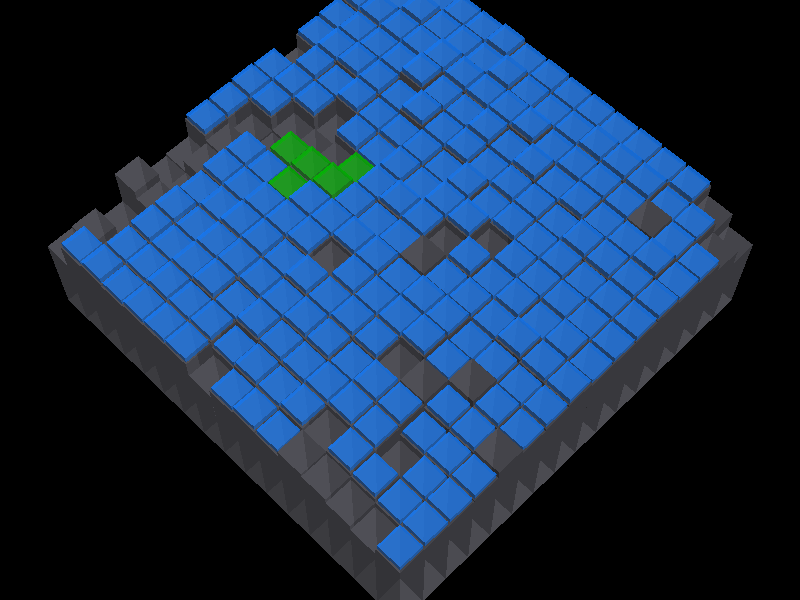}
\end{minipage}
\hfill
\begin{minipage}[t]{0.48\linewidth}
\centering
\includegraphics[width=\linewidth]{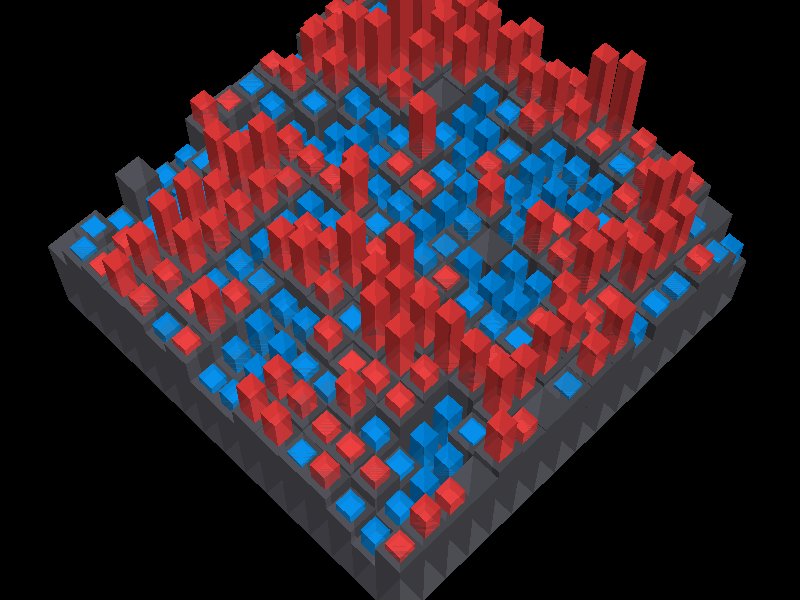}
\end{minipage}
\caption{Terrain Leveling (3, 4). (Left) Blue cells represent the new city after leveling. (Right) Terrain delta for the blast pattern. Red represents elevation decrease, blue represents elevation increase.}
\label{fig:figure-mb28-3-4}
\end{figure}

\paragraph{``Largest city'' algorithm}

A flood-fill algorithm finds the largest 4-connected region where neighbours stay within 0.2 units of any single neighbour.
A cliff may exist within a city provided a gentle slope connects both sides elsewhere.

\paragraph{Blast and rock simulation}

The charge is assumed sufficient to fracture rock down to the specified depth. The model does not need to calculate explosive size; a competent drill-and-blast team is assumed to handle this.

For each blast in the blast plan:

\begin{table}[H]
\centering\small
\begin{tabular}{@{}cp{0.88\linewidth}@{}}
\toprule
\textbf{Step} & \textbf{Action} \\
\midrule
1 & When detonation occurs at (x, y, depth), the heightmap is reduced by depth \\
2 & (Depth $\times$ 4) spheres in a vertical stack are added to a PyBullet sim at x,y \\
3 & The simulation runs until stable \\
4 & Where each sphere ended up, the heightmap there is raised pro-rata \\
5 & Rocks that fall off the world are lost \\
\bottomrule
\end{tabular}
\end{table}

Blasting the top of a mountain or flat plains with deeply drilled explosives has little effect, as the rocks mostly remain trapped in the resulting local minima and settle back to roughly the original shape. A more strategic approach is needed: the model must consider that rock rolls downhill and accumulates in valleys.

\paragraph{World generation}

Terrain is generated using Perlin noise with multiple octaves; the seed is deterministic per grid size and heights are normalized to $[0, 10]$. The world grid is $16{\times}16$ for subtask 0 and $48{\times}48$ for subtask 8.

\paragraph{Scoring}

Checks are performed in the following order:
\begin{enumerate}[nosep]
  \item Blast coordinates must lie within the grid.
  \item Cannot drill deeper than the current terrain height.
  \item Physics simulation (PyBullet): create heightfield collision shape, spawn spheres representing blasted material, simulate gravity and collision, track where spheres settle, and update the heightmap based on final sphere positions.
  \item Find the largest 4-connected region where all heights are within 0.2 units.
  \item Score $=$ (after $-$ before) / min(total\_cells/2, total\_cells $-$ before).
\end{enumerate}
Simulation results are cached between runs.

This task also tests model guardrails: refusing to discuss explosives yields an automatic score of 0, despite civil engineering and earthworks being safe and legal queries.

\paragraph{Verification}

A human-written reference solver places small charges (depth 1) at $z > 8$, sorted by $z$ so the sequence climbs from elevation 8 to the peaks, then repeats while gradually lowering the $z$ threshold until the median altitude is reached. Although this reblasts the same areas, it moves rock into valleys layer by layer.

\subsection{Interlocking Parts for 3D Printing}
\label{app:interlocking-parts}

\begin{promptbox}{Interlocking Parts for 3D Printing -- PROMPT}
I would like you to generate a cage of internal size 350 x 350 x 700 mm.

The cage should feature an all-around grid of 10mm thick bars, each with a square cross section. No thinner, no thicker.

The gap between bars should be between 5cm and 10cm, and uniform all over the model.

The cage needs to be split into parts that can be 3D printed on a 3D printer, with a build area of 400x400x50mm.

The cage assembly should be held together by 4 x M6 threaded rods threaded through the middle of the 4 vertical corner bars. You do not need to carve a female thread or nut slots into your parts, but should ensure that a rod fits with adequate clearance into every hole. Use no additional connectors. Do not include a door or lid.

Generate all parts that are needed for this project. They should be: - ready to slice - print without supports. - orientated such that they fit within the build volume - sitting on the z=0 plane. - not going to fall over during printing. - valid shapes - watertight, closed solids with no self intersections. - As few parts as possible. - Use as little build volume of the 3D printer as possible.

For each part file, you may provide an STL file, or OpenSCAD file that generates it, or a python script that generates either the OpenSCAD or the STL. Provide all parts in one go into the supplied output structure.

For each part, include a rigid transform matrix (4x4 row major) that shows where the part sits in 3D space when assembled, in your assembled 3D space, the cage sits on the z=0 plane and centred in x \& y around 0,0.

Take care to get this projection right as it will be used to test whether the parts fit together correctly.
\end{promptbox}

There is a single prompt, but 9 criteria are used to grade various aspects of the returned geometry.

The simplest valid solution requires 5 unique parts printed twice each (or 3 unique parts printed 2, 4, and 4 times), for a total of 10 parts. Most models produce incorrect designs, commonly assuming that $700/50 = 14$ parts suffice (which leaves unconnected vertical bars, as some slices lack a horizontal bar).

Many also assume that a single part printed 10 times is sufficient, which is easily disprovable visually:

\begin{figure}[ht]
\centering
\begin{minipage}[t]{0.48\linewidth}
\centering
\includegraphics[width=\linewidth]{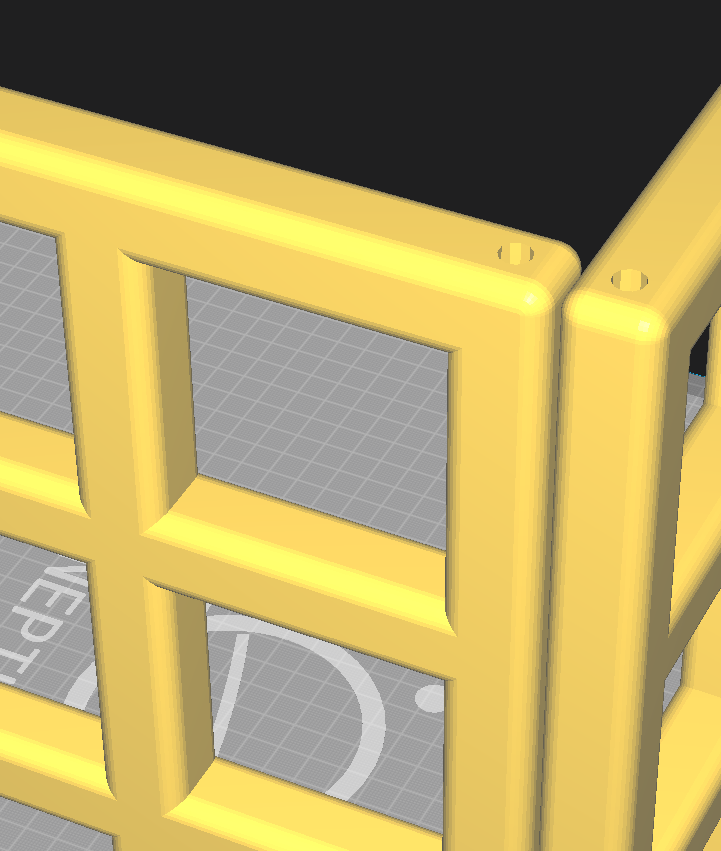}
\end{minipage}
\hfill
\begin{minipage}[t]{0.48\linewidth}
\centering
\includegraphics[width=\linewidth]{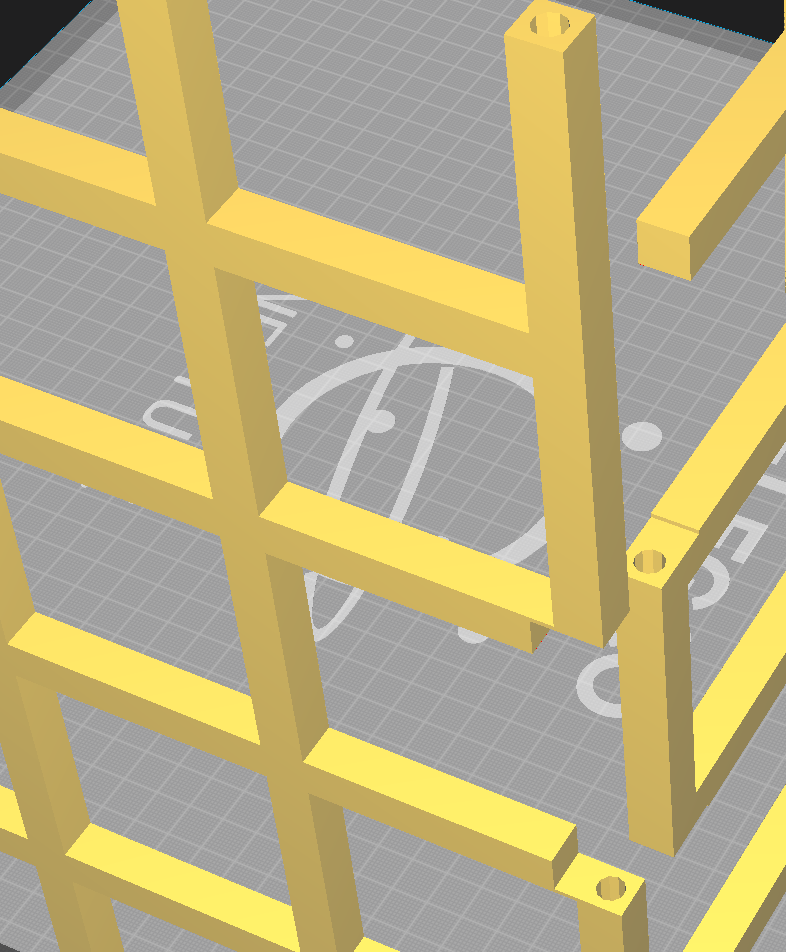}
\end{minipage}
\caption{Interlocking Parts (1, 2). (Left) A single threaded rod cannot hold these panels together, and the corner bar is twice as wide as the other bars. (Right) These panels interlock when oriented for assembly and can be held by a single threaded rod.}
\label{fig:figure-mb29-1-2}
\end{figure}

\noindent\textbf{Verification process:}

\begingroup
\small
\setlength{\LTleft}{0pt}
\setlength{\LTright}{0pt}
\begin{longtable}{@{}lp{0.78\linewidth}@{}}
\toprule
\textbf{Subpass} & \textbf{Check} \\
\midrule
\endfirsthead
\toprule
\textbf{Subpass} & \textbf{Check} \\
\midrule
\endhead
\bottomrule
\endfoot
\bottomrule
\endlastfoot
0: Basic validation & Exactly 10 parts; each part is or generates valid STL; non-zero volume; watertight meshes; consistent face winding; no non-manifold geometry; single connected body per part \\
1: Printability & Parts fit in build volume ($400{\times}400{\times}50$); rest on $z{=}0$ plane; sufficient bed contact area ($\geq 500$\,mm\textsuperscript{2}); no large unsupported overhangs; will not fall over during printing \\
2: Bar gap verification & 100 randomly placed cubes, per part, are used to check that: 40mm cubes can pass through gaps (gap $\geq$ 50mm) due to at least one intersection test resulting in a zero volume per part; 101mm cubes cannot pass through (gap $\leq$ 100mm), as all intersections must have a positive volume \\
3: Assembly size & Assembled cage fits 350$\times$350$\times$700 internal space; not larger than 380$\times$380$\times$740 external \\
4: Rod clearance & 6mm rod passes through corners without intersection; 16mm rod blocked (adequate material around holes) \\
5: Part interference & No two parts intersect when assembled. All 90 combinations are checked \\
6: Rod engagement & Each part contacts at least 2 rods (a single rod contact would form a door or hinge, which is banned in the prompt) \\
7: Bar thickness & Ray-based thickness measurement; dominant thickness \textasciitilde{}10mm \\
8: Gap uniformity & Ray-based gap measurement; dominant gap in 50--100mm range \\
\end{longtable}
\endgroup

Scoring: pass/fail per subtask; partial credit based on subtasks passed.

\begin{figure}[ht]
\centering
\includegraphics[width=0.55\linewidth]{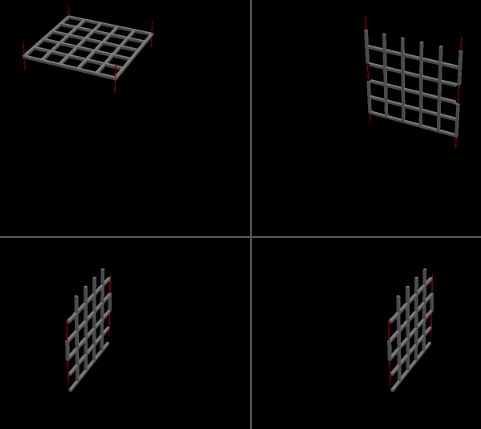}
\caption{Interlocking Parts (3). Rod engagement tests performed using OpenSCAD (subtask 6). Zero rods result in a panel falling out; one rod results in a hinge.}
\label{fig:figure-mb29-3}
\end{figure}

\paragraph{Solvability:}

A human designed the parts in OpenSCAD and calculated the assembly matrix by hand; this reference solution passes all tests.

\begin{figure}[ht]
\centering
\begin{minipage}[t]{0.48\linewidth}
\centering
\includegraphics[width=\linewidth]{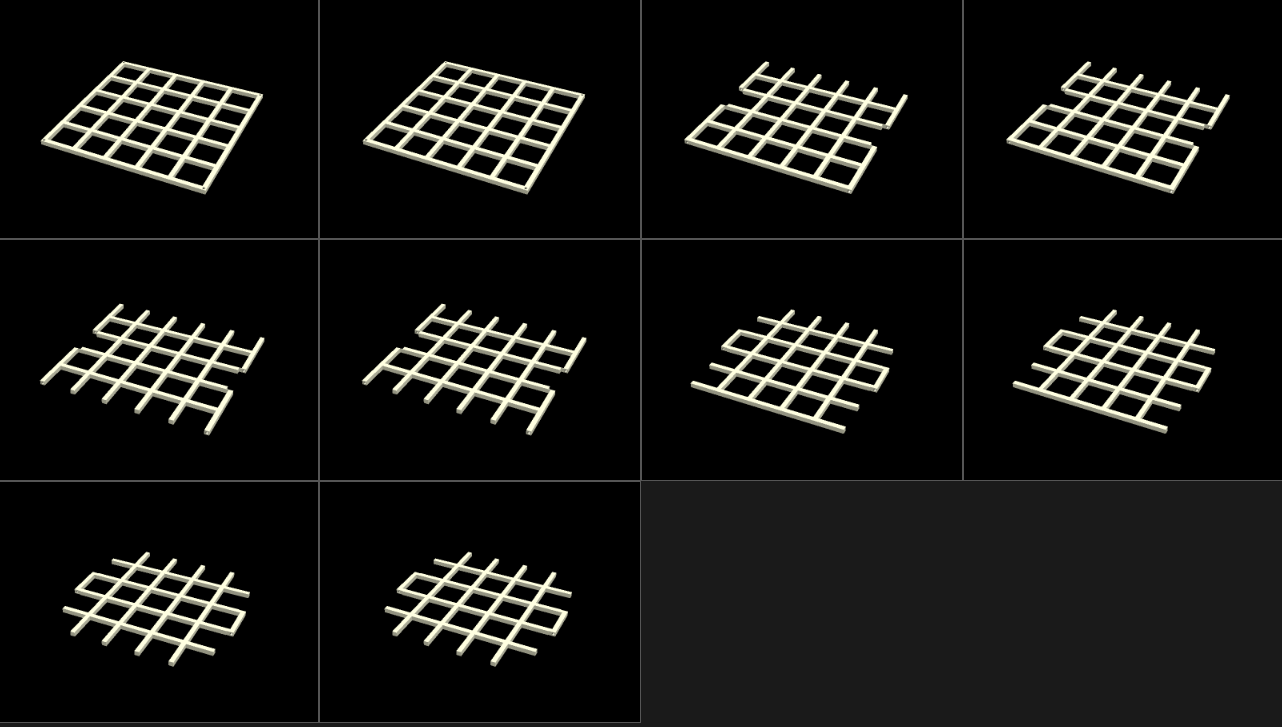}
\end{minipage}
\hfill
\begin{minipage}[t]{0.48\linewidth}
\centering
\includegraphics[width=\linewidth]{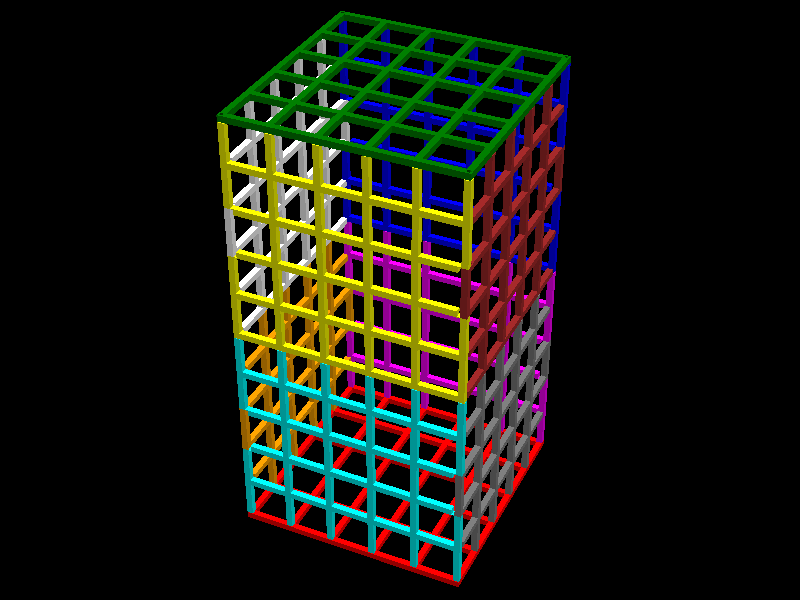}
\end{minipage}
\caption{Interlocking Parts (4). A valid (human-created) solution, showing all 10 parts and the completed assembly.}
\end{figure}

\subsection{Largest 3D-Printable Prime Number}
\label{app:largest-prime}

Construct the largest prime number that can be 3D-printed without support material. Without additional constraints the answer is unbounded, so we stipulate that no triplet (three digits between commas in decimal notation) may be repeated (e.g.\ 123{,}123 is banned; 123{,}000{,}001{,}230 is permitted).

A 7-segment font is used, and digits may be rotated or flipped as needed, including off-axis. Structured output with a schema extracts the orientation and digit array:

\begin{outputbox}{Largest 3D-Printable Prime -- OUTPUT SCHEMA}
\{ "type": "object", "properties": \{

~~"numberSequence": \{ "type": "array", "items": \{

~~~~"type": "object", "properties": \{

~~~~~~"digit": \{"type": "integer"\},

~~~~~~"orientation": \{"type": "string",

~~~~~~~~"enum": ["flat","flippedX","flippedY",

~~~~~~~~~~~~~~~~~~"rotate90X","rotate90Y","rotate180Z"]\}

~~~~\} \} \} \} \}
\end{outputbox}

Orientations are only included in the enum if there exists a digit that can be printed in that orientation successfully.

\paragraph{Solution insights}

This task has two components. The mathematical problem is computationally demanding: finding a prime with no repeated three-digit triplet yields an answer approximately 3{,}000 digits long.

The visual problem is simpler: each digit and orientation choice constrains future choices, and the search space shrinks rapidly. For example, the digit 8:

\begin{table}[H]
\centering\small
\begin{tabular}{@{}l@{}}
\toprule
\textbf{Digit 8 constraints} \\
\midrule
Cannot be printed on its long or short side (bridge overhangs) \\
When printed flat, must not occur after any number except 8 (or the start) \\
\bottomrule
\end{tabular}
\end{table}

This single observation causes the possibilities to diverge into 5 branches and shrinks the overall search space by {\raise.17ex\hbox{$\scriptstyle\sim$}}10\%:

\begin{table}[H]
\centering\small
\begin{tabular}{@{}l@{}}
\toprule
\textbf{Regex branch} \\
\midrule
\texttt{\textasciicircum{}[012345679,]+} \\
\texttt{\textasciicircum{}8[012345679,]+} \\
\texttt{\textasciicircum{}88[012345679,]+} \\
\texttt{\textasciicircum{}888,[012345679,]+} \\
\texttt{\textasciicircum{}8,888[012345679,]+} \\
\texttt{\textasciicircum{}88,888[012345679,]+} \\
\bottomrule
\end{tabular}
\end{table}

Insights like this narrow the solution space considerably. The model is asked to prioritise number size over the tallest stack, but to aim for a height of at least 70\,mm. This constraint was added because several models discovered the ``9 can be printed on 8 but not 8 on 9'' rule.

The tallest and largest reference numbers differ:
\begin{itemize}[nosep]
  \item \textbf{Tallest:} 88{,}888{,}999{,}996{,}969{,}699{,}696{,}669{,}666{,}333{,}337{,}773{,}111 (the trailing 1s form a 9\,cm spire balanced on a rotated 3)
  \item \textbf{Largest:} 888{,}999{,}996{,}969{,}699{,}696{,}669{,}666{,}333{,}377{,}777{,}711{,}111{,}171
\end{itemize}

\begin{figure}[ht]
\centering
\begin{minipage}[t]{0.48\linewidth}
\centering
\includegraphics[width=\linewidth]{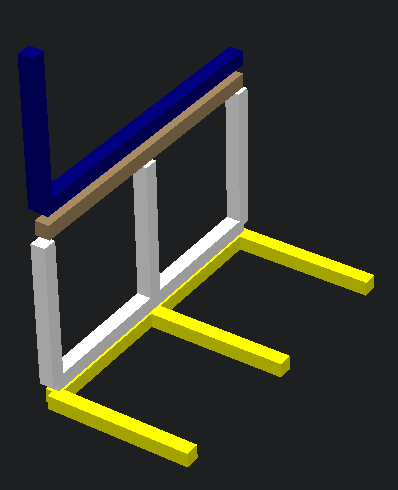}
\end{minipage}
\hfill
\begin{minipage}[t]{0.48\linewidth}
\centering
\includegraphics[width=\linewidth]{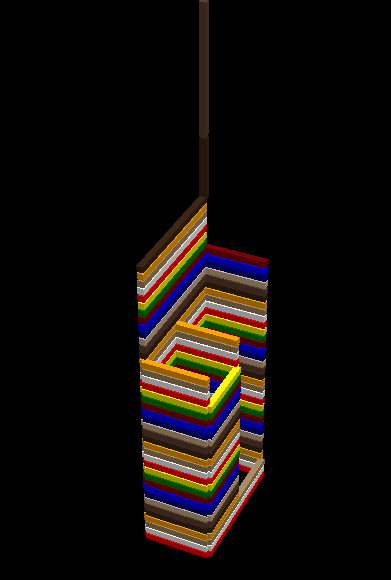}
\end{minipage}
\caption{Largest 3D-Printable Prime (1, 2). (Left) A model's suggestion of 3{,}317 is not 3D-printable: the brown digit 1 sags without support, scoring 0. (Right) The largest 3D-printable prime without repeating triplets.}
\label{fig:figure-mb30-1-2}
\end{figure}

\noindent\textbf{Verifier mechanics:}

\begin{table}[htb]
\centering\small
\begin{tabular}{@{}cp{0.82\linewidth}@{}}
\toprule
\textbf{Step} & \textbf{Check} \\
\midrule
1 & \textbf{Triplet constraint.} No sequence of 3 consecutive digits repeated (constrains the search space to prevent infinite solutions) \\
2 & \textbf{Printability.} Check for overhangs in each digit's orientation. 8 cannot be printed on its long or short side; 7 can be printed on its head and side but not its base; 3 can only go on its long side \\
3 & \textbf{Stackability.} Flat-on-flat: determined by which segments are in use. 7 can go on 3, but not vice versa. 6 cannot go on 9 flat, but can when rotated 180\textdegree{} or flipped. Only 7s, 3s, and 1s can go non-flat. Only vertical 1 ``spires'' can go on top of non-flat 3s or 7s \\
4 & \textbf{Height calculation.} Flat: $+1.2$\,mm; rotate90X: $+21.2$\,mm; rotate90Y: $+11.2$\,mm (or $1.2$\,mm for ``1''). A 0.2\,mm gap in $Z$ produces parts that hold together during printing but snap apart afterwards \\
5 & \textbf{Primality test} (Baillie--PSW). Miller--Rabin base 2; strong Lucas test; results cached \\
6 & \textbf{Scoring.} Score $=$ len(answer) / len(known\_best); known best: 42-digit number; 10\% penalty if height ${<}\,70$\,mm \\
\bottomrule
\end{tabular}
\end{table}

\paragraph{Solvability}

A human-written solver splits the problem into two parts: flat digits and rotated digits.
Flat digits are arranged first. 8s must go first, then interleaved 6s and 9s, then 3s, then 7s. If any 2 or 5 appears then 6 or 9 cannot, and 9 is larger than 5. If any 0 appears then 9 cannot, and 9 is larger than 0.
There are 29 possible prefix combinations (comma position is unknown until the suffix is calculated):

\begin{verbatim}
for i8 in range(3, 6):
  for i3 in range(3, 6):
    for i7 in range(3, 6):
      prefixes.append(
        int("8" * i8 + (
          "999""996""969""699""696""669""666") +
          "3" * i3 + "7" * i7))
\end{verbatim}

The tree of valid suffixes is then computed by solving the stacking constraints for 1s, 3s, and 7s. The suffix begins after a flat 7, so any 3 must be rotated.

\begin{verbatim}
for suffix in itertools.product(
    ["", "3", "1", "7"], repeat=10):
  suffix = "".join(suffix)
  topBarAllowed = True
  longSideAllowed = True

  invalid = False
  for s in suffix:
    if topBarAllowed and longSideAllowed:
      if s == "7": continue
      if s == "3":
        # We have to rotate the 3
        topBarAllowed = False
        longSideAllowed = False
      if s == "1":
        # We lose ability to print top bar.
        topBarAllowed = False
    elif longSideAllowed:
      if s == "1": continue
      if s in ["3", "7"]:
        # Must rotate 3 or 7 spikes up
        topBarAllowed = False
        longSideAllowed = False
    else:
      if s == "1": continue
      invalid = True

  if invalid:
    continue

  suffixes.append(suffix)
\end{verbatim}

The product of prefixes and suffixes is then iterated:

\begin{table}[H]
\centering\small
\begin{tabular}{@{}cp{0.88\linewidth}@{}}
\toprule
\textbf{Step} & \textbf{Action} \\
\midrule
1 & Skip if smaller than the current best \\
2 & With comma positions now known, check for repeated triplets \\
3 & Run primality test on the {\raise.17ex\hbox{$\scriptstyle\sim$}}40-digit candidate \\
4 & Record as new best; continue searching \\
\bottomrule
\end{tabular}
\end{table}

Only about 50 primality tests are needed; the search completes in a few minutes.

\subsection{Topology Enumeration}
\label{app:topology-enumeration}

Given a unit square with labeled corners, enumerate all label configurations that force class interfaces (where different classes must meet).
See the figures below for a visual explanation of the task:

\begin{figure}[ht]
\centering
\begin{minipage}[t]{0.48\linewidth}
\centering
\includegraphics[width=\linewidth]{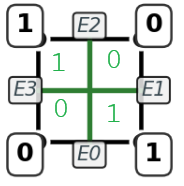}
\end{minipage}
\hfill
\begin{minipage}[t]{0.48\linewidth}
\centering
\includegraphics[width=\linewidth]{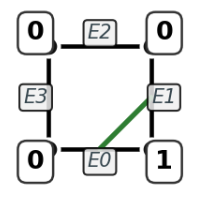}
\end{minipage}
\caption{Topology Enumeration (1, 2). (Left) Labelling the 4 corners with alternating 1s and 0s ensures that a border between classes must touch all edges, and thus a boundary between the two classes must exist inside the square. (Right) 3 corners labeled 0 and 1 labeled 1 infers a class boundary intersecting 2 edges, and thus there are two regions inside the square.}
\label{fig:figure-vbg1-1-2}
\end{figure}

\begin{promptbox}{Topology Enumeration -- EXACT PROMPT}
You are given a unit square with corners ordered (bottom-left, bottom-right, top-right, top-left). Each corner is labeled from \{0, 1, 2...\}. Boundaries inside may be any continuous curves; only corner labels are observed.

Assume exactly 2 distinct classes occur anywhere in or on the square.

List all corner-label configurations (4-tuples, in the order above) that are sufficient to guarantee that 2 distinct classes meet somewhere inside the square. Canonicalisation: relabel by first occurrence (scan left-to-right; first new label -> 0, next -> 1, ...). Treat any label renamings as identical; list each equivalence class once.

\end{promptbox}

\begin{outputbox}{Topology Enumeration -- OUTPUT SCHEMA}
\{ "type": "object", "properties": \{

~~"configs": \{ "type": "array", "items": \{

~~~~"type": "array", "minItems": 4, "maxItems": 4,

~~~~"items": \{"type": "integer"\}

~~\} \}

\}, "required": ["configs"] \}
\end{outputbox}

Multiple subpasses exist, these shuffle the required order that the output labels must be provided and the number of classes.
A ground truth set was calculated by the test author, and scoring is binary against that.

\subsection{Topology Edge Tasks: Enumerate Edges}
\label{app:enumerate-edges}

Given corner labels and ordering, enumerate all edges guaranteed to exist between adjacent corners.

\begin{figure}[ht]
\centering
\includegraphics[width=0.55\linewidth]{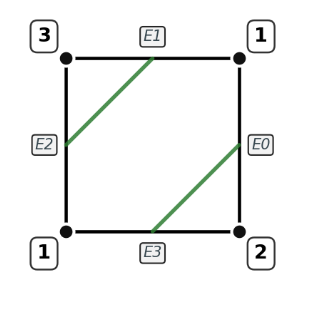}
\caption{Enumerate Edges (1). Given the vertex labels, boundaries between regions are unavoidable. List all the connections that are guaranteed to exist (green lines).}
\label{fig:figure-vgb2-1}
\end{figure}

\begin{promptbox}{Enumerate Edges -- TYPICAL PROMPT}
Squares (each tuple lists the four corner labels; integers denote distinct classes): (2, 1, 1, 1) (2, 1, 2, 1) (2, 1, 3, 1) (2, 2, 1, 1) (2, 2, 2, 1)

You are given unit squares with corners labelled in ('bottom-right', 'top-right', 'top-left', 'bottom-left') order. Edges are indexed: right=0, top=1, left=2, bottom=3.

For each square above (in the same order), list which edges are guaranteed to connect. Return a list where each element is a list of sorted [i,j] pairs (i < j). If no edges are deterministically guaranteed (including ambiguous cases), return [] for that square.
\end{promptbox}

\begin{outputbox}{Enumerate Edges -- TYPICAL STRUCTURED OUTPUT}
\{'edges': [[[1, 2],[0,3]], [], [], [], [[0, 1]]]\}
\end{outputbox}

Ground truth was calculated by the test author, and grading is binary against it. Note that multiple squares are tested per prompt.

\subsection{Topology Edge Tasks: Classify Behaviour}
\label{app:classify-behaviour}

\begin{figure}[ht]
\centering
\includegraphics[width=0.55\linewidth]{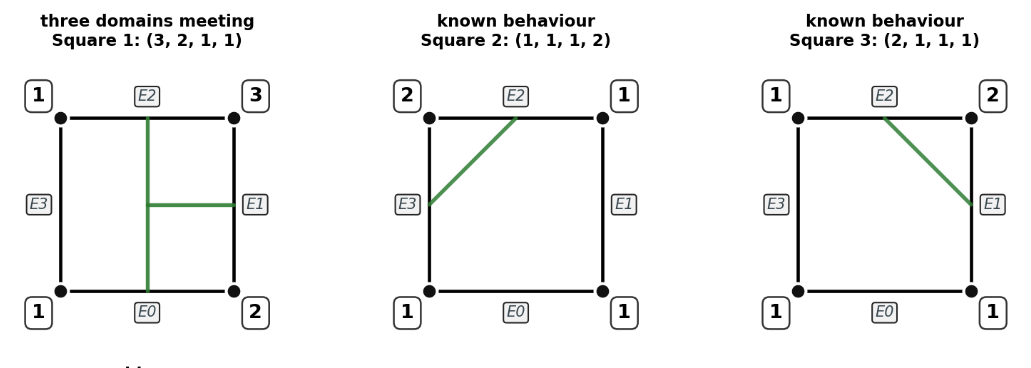}
\caption{Classify Behaviour (1). Classifying the intersections.}
\label{fig:figure-vgb3}
\end{figure}

Classify edge behaviours in labeled squares (known behaviour, three domains meeting, ambiguous).

\begin{promptbox}{Classify Behaviour -- TYPICAL PROMPT}
Squares (each tuple lists the four corner labels; integers denote distinct classes): (3, 2, 1, 1) (1, 1, 1, 2) (2, 1, 1, 1) (3, 2, 1, 4)

You are given unit squares with corners labelled in ('top-right', 'bottom-right', 'bottom-left', 'top-left') order.

Classify each square's topological behaviour (in the same order) as one of: 'known behaviour', 'three domains meeting', or 'ambiguous'. Definitions: 'known behaviour' = deterministic edge connections can be made; 'three domains meeting' = deterministic edge connections where three distinct classes meet at a point; 'ambiguous' = multiple valid topologies could exist. Return a list of exact label strings.
\end{promptbox}

\begin{outputbox}{Classify Behaviour -- STRUCTURED OUTPUT SCHEMA}
\{ "type": "object", "properties": \{ "labels": \{ "type": "array", "items": \{ "type": "string", "enum": ["known behaviour", "three domains meeting", "ambiguous"] \} \} \}, "required": ["labels"] \}
\end{outputbox}

Verifier requires a binary exact match to answers calculated by the test author.

\subsection{Half Subdivision Neighbours}
\label{app:half-subdivision-neighbours}

In a recursively bisected unit square/cube, identify all neighbours of a target cell.

\begin{promptbox}{Half Subdivision Neighbours -- TYPICAL PROMPT}
You are given a binary tree describing an axis-aligned half subdivision of the unit square.

Each node splits its parent cell into two children by bisecting along axes in the repeating cycle x $\rightarrow$ y $\rightarrow$ y $\rightarrow$ x $\rightarrow$ x (repeating).

Here is the subdivision tree:

\begin{verbatim}
""  
|---- 0  
|   |---- 00  
|   |   |---- 000  
|   |   |   |---- 0000  
|   |   |   |   |---- 00000  
|   |   |   |   |   |---- 000000  
|   |   |   |   |   |   |---- 0000000  
|   |   |   |   |   |   |   |---- 00000000  
|   |   |   |   |   |   |   `---- 00000001  
|   |   |   |   |   |   `---- 0000001  
|   |   |   |   |   |       |---- 00000010  
|   |   |   |   |   |       `---- 00000011  
|   |   |   |   |   `---- 000001  
|   |   |   |   |       |---- 0000010  
|   |   |   |   |       |   |---- 00000100  
|   |   |   |   |       |   `---- 00000101  
|   |   |   |   |       `---- 0000011  
|   |   |   |   |           |---- 00000110  
|   |   |   |   |           `---- 00000111
\end{verbatim}

(\textasciitilde{}200 lines trimmed for brevity)

Target leaf: 01001000
\end{promptbox}

\begin{outputbox}{Half-Subdivision Neighbours -- TYPICAL STRUCTURED OUTPUT}
\{'neighbors': [ '01000100', '01001100', '01001001', '00101011']\}
\end{outputbox}

Neighbours specified as binary strings encoding the subdivision path.
The answers are precalculated by a generator and set arithmetic is used to grade membership of the group, as shown below:

\begin{figure}[ht]
\centering
\includegraphics[width=0.85\linewidth]{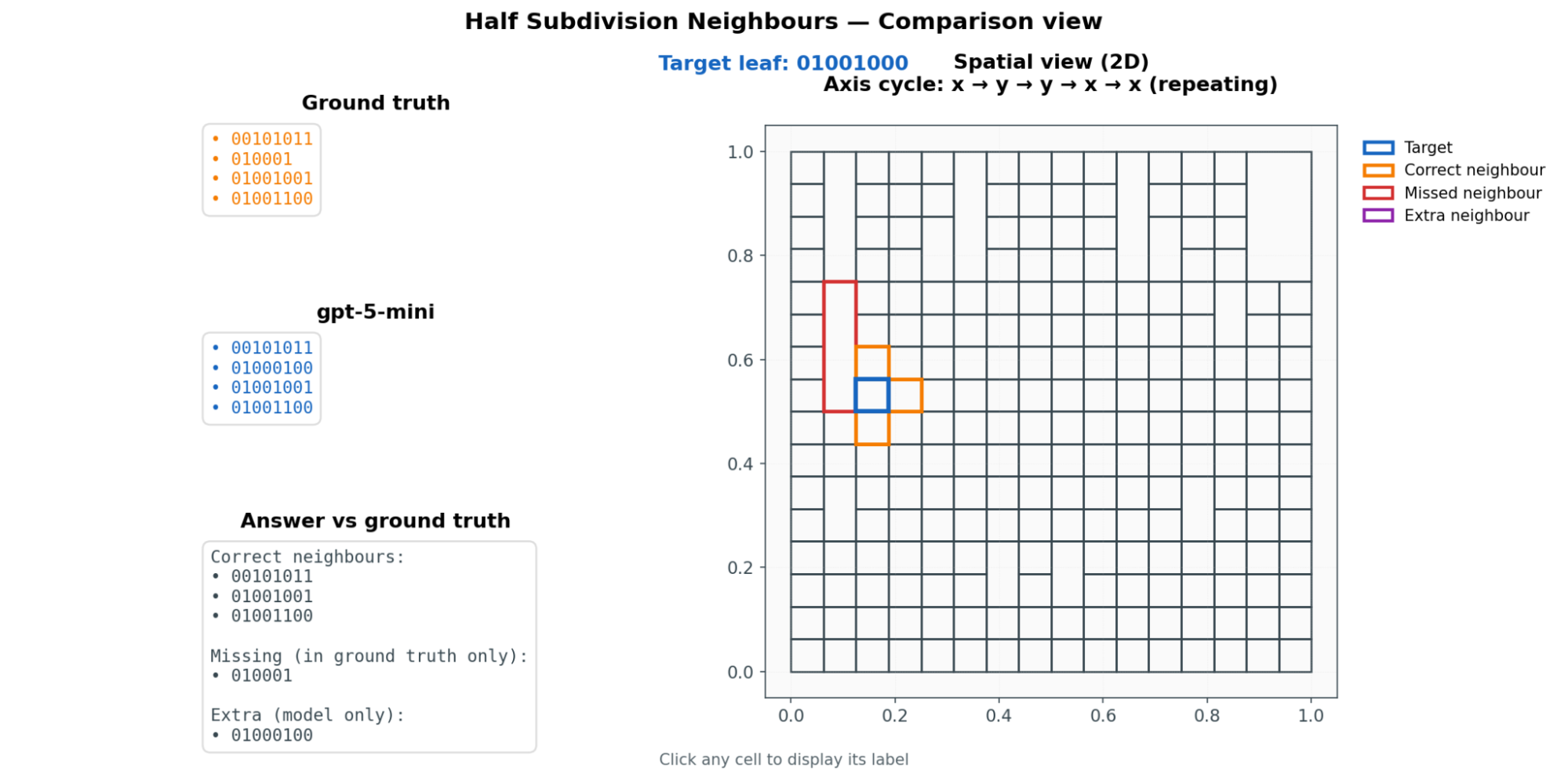}
\end{figure}

\paragraph{Subpass variations}

The test varies from 2D to 3D, with different tree depths, and by varying the split dimension order in the binary format.
Examples of binary split formats:

\begin{table}[H]
\centering\small
\begin{tabular}{@{}l@{}}
\toprule
\textbf{Split cycle} \\
\midrule
Z, Y, X, Z, Y, X \ldots{} \\
X, Y, Z, X, Y, Z \ldots{} \\
X, X, Y, Z, X, X, Y, Z \ldots{} \\
Z, Y, X, X, Z, Z, Y, X, Z \ldots{} \\
Y, Y, X, Y, Z, Y, Y, X, Y, Z \\
\bottomrule
\end{tabular}
\end{table}

The figure below shows an example of a 3D split. A model's attempt missed the purple-shaded cells:

\begin{figure}[ht]
\centering
\includegraphics[width=0.55\linewidth]{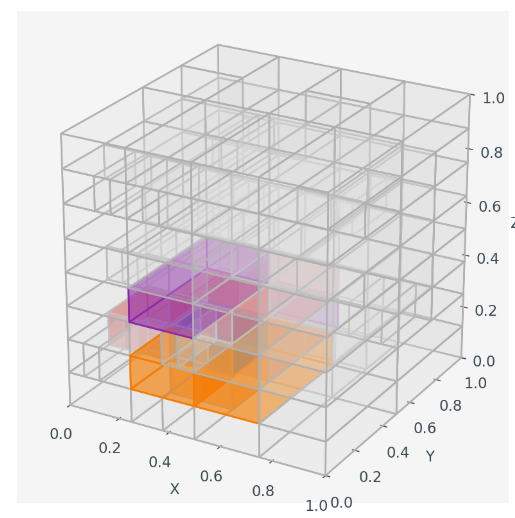}
\caption{Half Subdivision Neighbours (2). Subdivision neighbour test performed in 3D.}
\label{fig:figure-vgb5-2}
\end{figure}

\subsection{Delaunay Triangulation}
\label{app:delaunay-triangulation}

\begin{promptbox}{Delaunay Triangulation -- TYPICAL PROMPT}
You are given a set of 2D points in general position (indices correspond to the order shown): [ [0.444, 0.568], [0.908, 0.254], [0.589, 0.359], [0.756, 0.543], [0.202, 0.516], [0.242, 0.05], [0.113, 0.343], [0.015, 0.773] ]

Return the Delaunay triangulation as a list of triangles. Each triangle is a list of three point indices (sorted in ascending order).
\end{promptbox}

\begin{outputbox}{Delaunay Triangulation -- TYPICAL OUTPUT}
\{'triangles': [ [0, 1, 2], [0, 1, 5], [0, 3, 5], [1, 2, 5], [2, 4, 5], [3, 4, 5]]\}
\end{outputbox}

As can be shown from the figure below, the [0,1,2] triangle is an error.

\begin{figure}[ht]
\centering
\includegraphics[width=0.55\linewidth]{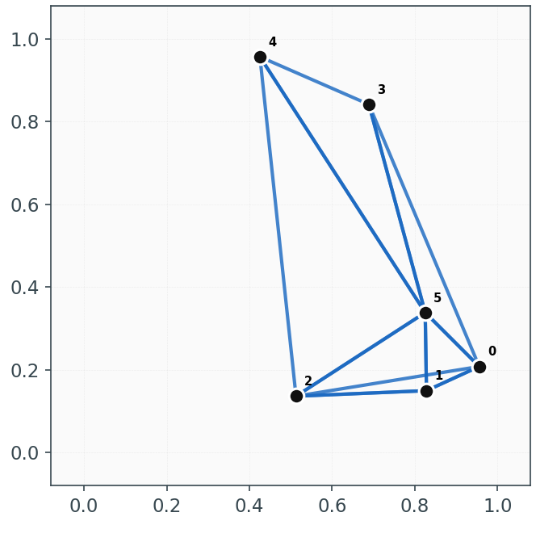}
\caption{Delaunay Triangulation (1). This triangulation required 5 triangles; 6 were provided.}
\label{fig:figure-vgb6-1}
\end{figure}

\noindent\textbf{Verifier strategy:}

\begin{table}[H]
\centering\small
\begin{tabular}{@{}l@{}}
\toprule
\textbf{Check} \\
\midrule
All triangles use valid point indices \\
Triangulation is complete (covers convex hull) \\
Delaunay criterion: no point inside circumcircle of any triangle \\
This can't use the ``ground truth'' reference without first confirming there are no co-circular points \\
\bottomrule
\end{tabular}
\end{table}

\subsection{Shikaku Rectangles}
\label{app:shikaku-rectangles}

\textbf{Objective:} Solve a Shikaku instance by partitioning a rectangular grid into axis-aligned rectangles such that each rectangle contains exactly one clue number, and the rectangle's area equals that clue.

\textbf{Input:} The prompt provides a $H \times W$ grid with some cells containing integers (the clues). Empty cells have no number.

\textbf{Output:} Return a list of rectangles, each encoded as a 4-tuple of integer coordinates $[x_0, y_0, x_1, y_1]$ (bounding-box corners), covering the whole grid with no overlaps or gaps.

\begin{figure}[H]
\centering
\includegraphics[width=0.45\linewidth]{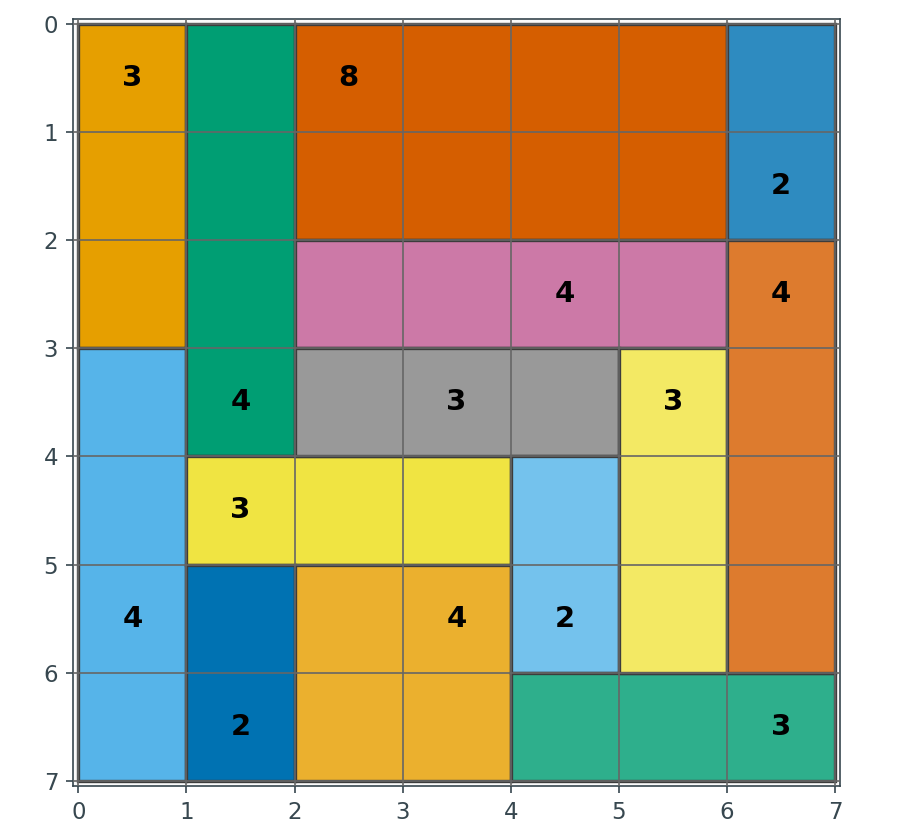}
\caption{Shikaku Rectangles. A solved $7{\times}7$ instance: each coloured rectangle contains exactly one clue equal to its area.}
\label{fig:shikaku-example}
\end{figure}

\begin{outputbox}{Shikaku Rectangles -- OUTPUT SCHEMA}
\{ "type": "object", "properties": \{

~~"rectangles": \{ "type": "array", "items": \{

~~~~"type": "array", "minItems": 4, "maxItems": 4,

~~~~"items": \{"type": "integer", "minimum": 0\}

~~\} \}

\}, "required": ["rectangles"] \}

Each rectangle: [x\_min, y\_min, x\_max, y\_max]
\end{outputbox}

\textbf{Verifier:}

\begin{table}[H]
\centering\small
\begin{tabular}{@{}l@{}}
\toprule
\textbf{Check} \\
\midrule
Rectangles tile the grid exactly (no gaps, no overlaps) \\
Each rectangle contains exactly one clue number \\
Each rectangle's area equals its contained clue \\
\bottomrule
\end{tabular}
\end{table}

\subsection{Two Segments}
\label{app:two-segments}

\textbf{Objective:} Place two line segments on the boundary of a unit square such that they partition the interior into a target number of polygonal regions (e.g.\ triangles, quadrilaterals, pentagons).

\textbf{Input:} The prompt specifies the square and the target region count and types that the two segments must produce.

\textbf{Output:} Two segments as endpoint coordinate pairs, returned via structured output.

\textbf{Verification:} A deterministic geometric verifier computes the regions induced by the two segments and the square boundary, then checks that the resulting partition matches the target region count and polygon types.

\begin{figure}[H]
\centering
\includegraphics[width=0.4\linewidth]{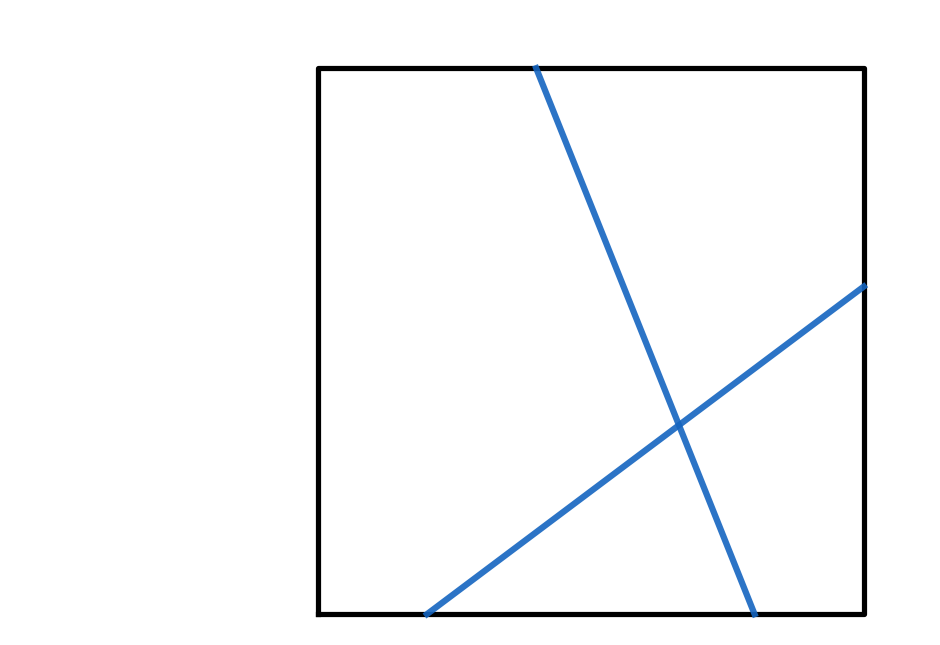}
\caption{Two Segments. The model must place two segments on the square boundary to partition the interior into a target set of regions (here: 1 pentagon, 2 quadrilaterals, and 1 triangle).}
\label{fig:two-segments-example}
\end{figure}

\begin{outputbox}{Two Segments -- OUTPUT SCHEMA}
\{ "type": "object", "properties": \{

~~"segments": \{ "type": "array", "minItems": 2, "maxItems": 2, "items": \{

~~~~"type": "array", "minItems": 2, "maxItems": 2, "items": \{

~~~~~~"type": "array", "minItems": 2, "maxItems": 2,

~~~~~~"items": \{"type": "number"\}

~~~~\} \} \}

\}, "required": ["segments"] \}
\end{outputbox}

\textbf{Verifier:} The induced partition is computed and compared against the target region count and polygon types; endpoint positions are checked with a geometric tolerance.

\end{document}